\documentclass[10pt,twocolumn,letterpaper]{article}

\usepackage{cvpr}
\usepackage{times}
\usepackage{epsfig}
\usepackage{graphicx}
\usepackage{amsmath}
\usepackage{amssymb}
\usepackage{amscd}
\usepackage{amsthm}

\usepackage{multirow}
\usepackage{enumitem}
\usepackage[]{algorithm2e}
\usepackage{subcaption}
\usepackage{flushend}

\def\mat#1{\mathchoice{\mbox{\bf$\displaystyle\bf#1$}}
	{\mbox{\bf$\textstyle\bf#1$}}
	{\mbox{\bf$\scriptstyle\bf#1$}}
	{\mbox{\bf$\scriptscriptstyle\bf#1$}}}
\def\m#1{\protect\mat #1}

\usepackage[pagebackref=true,breaklinks=true,letterpaper=true,colorlinks,bookmarks=false]{hyperref}

\cvprfinalcopy 

\def\cvprPaperID{1309} 

\begin{document}

\title{Robust Optical Flow Estimation of Double-Layer Images \\ under Transparency or Reflection}

\author{Jiaolong Yang\thanks{J. Yang is with Beijing Lab of Intelligent Information Technology, Beijing Institute of Technology (BIT), and with College of Engineering and Computer Science, the Australian National University (ANU). \mbox{Email: jiaolong.yang@anu.edu.au}}\\
BIT \& ANU\\
\and
Hongdong Li\\
ANU \& ACRV\\
\and
Yuchao Dai\\
CECS, ANU\\
\and
Robby T. Tan\\
Yale-NUS College\\
}

\maketitle
\begin{abstract}
This paper deals with a challenging, frequently encountered, yet not properly investigated problem in two-frame optical flow estimation. That is, the input frames are compounds of two imaging layers -- one desired background layer of the scene, and one distracting, possibly moving layer due to transparency or reflection. In this situation, the conventional brightness constancy constraint -- the cornerstone of most existing optical flow methods -- will no longer be valid. In this paper, we propose a robust solution to this problem. The proposed method performs both optical flow estimation, and image layer separation. It exploits a generalized double-layer brightness consistency constraint connecting these two tasks, and utilizes the priors for both of them. Experiments on both synthetic data and real images have confirmed the efficacy of the proposed method. To the best of our knowledge, this is the first attempt towards handling generic optical flow fields of two-frame images containing transparency or reflection.
\end{abstract}
\section{Introduction}
Most optical flow methods assume that there is only one imaging layer on the observed image with the brightness of scene objects, and use the brightness constancy constraint (BCC) to estimate the optical flow for scene objects. This single imaging layer assumption, however, can be often violated in real-world situations, especially in cases involving transparency or reflection.  Transparencies and reflections are frequently met in imaging process, \eg, when one is looking at street scene from inside a car through a stained windscreen, or seeing through a thin layer of rain, looking into a window with semi-reflections on the window surface \etc. The BCC will generally not hold for the resultant double-layer images, even in ideal noise-free cases.

In all the above examples, the observed image $\m I$ can be modeled as a superposition of two constituting layers, denoted as $\m I =\m L_1\oplus\m L_2$, where $\oplus$ denotes some suitable layer combination operator.  Without loss of generality, we call $\m L_1$ the background scene layer, which corresponds to the image of the desired scene that we intend to capture, and $\m L_2$ the foreground distracting layer, which corresponds to the semi-transparent media (\eg a glass window with dirt or reflections on it) or the semi-reflected image.

The main goal of this paper is to robustly estimate the optical flow field of the scene objects (\ie the background layer), which is of concern for vision systems. We consider two general cases: the foreground distracting layer is \emph{stationary}, or \emph{dynamically changing}.

Let $\m I$ and $\m I'$ be two time-consecutive frames of a scene containing the aforementioned two layers. In the presence of a dynamic foreground layer, there are two legitimate optical flow fields -- one for the foreground layer and one for the background layer. Denote the two flow fields generated by the movements of the two layers as $\m U$ and $\m V$ respectively. The relationships among the observed images, the image layers and the optical flow fields can be given as
\begin{center}
$\begin{CD}
\m I~=~\m L_1~~@.{\oplus} @.~~~\m L_2\\
~~~~~~@VV{\m U }V~~~ @.   ~@VV{\m V }V \\
\m I'~=~\m L_1'~~@.{\oplus} @.~~~\m L_2'\\
\end{CD}$\\
\end{center}
When $\m V\equiv\m 0$ and $\m L_2\equiv\m L_2'$, \ie the foreground layer is static, our task is to estimate a single flow field $\m U$ for background layer, and also estimate the layers $\m L_1$, $\m L_1'$, $\m L_2$. Otherwise when a dynamic foreground layer exists, we will estimate two flow fields $\m U, \m V$ as well as the layers $\m L_1$, $\m L_1'$, $\m L_2$, $\m L_2'$. As we explicitly perform image layer separation (\ie estimating $\m L_1$, $\m L_1'$, $\m L_2$, $\m L_2'$), an appealing byproduct of our method is the restoration of the clear scene images.

For either of the two cases with a static or dynamic foreground layer, this is a highly ill-posed problem, especially considering optical flow estimation and image layer separation problems \emph{per se} are known to be ill-posed. From only two input images, our task is to recover one or two optical fields, as well as the two unknown layers.

Little work has been reported in the literature concerning this double-layer image optical flow estimation problem, with only a few exceptions in the early days of computer vision research, \eg \cite{shizawa1990simultaneous}\cite{shizawa1991unified}\cite{langley1992transparent}\cite{darrell1993nulling}.  These works however often used over-simplified assumption and restrictive motion field models, such as assuming a constant flow field over time or over space (\eg. globally translating). Bergen \etal~\cite{bergen1992three} proposed a ``three-frame algorithm" to recover two constituting flow fields, assuming the flow field is constant over at least three frames.

In contrast, this paper removes these restrictive assumptions, and proposes a two-frame algorithm for robustly recovering the flow field(s).  Our method works for generic motions, and is thus applicable to a much wider range of practical situations for robust optical flow estimation.

\subsection{Related Works}
This paper is concerned with optical flow estimation in double-layer images where both layers can possibly be moving. Despite that the phenomena of such multiple imaging layers and motions are frequently encountered in reality, few papers in the literature have been devoted to this topic.  This is in a sharp contrast to the existence of vast amount of papers on the classic optical flow problems (an analysis of recent practices of optical flow can be found in \cite{Sun_review}).

One of the first work for multiple optical flow computation is possibly due to Shiwaza \etal\cite{shizawa1990simultaneous}\cite{shizawa1991unified}.  By assuming the two underlying flow fields to be constant (\eg pure translating), they derived a generalized brightness constancy constraint for the multi-motion case.  However, this constant motion assumption is restrictive, not applicable for general flow fields with complex motions. Nevertheless, their method, being one of the first, has inspired a number of variants and extensions \cite{pingault2002optical}\cite{auvray2009jointmotion}\cite{ramirez2006multi}\cite{toro2000multiple}.  Some variants operate in the Fourier domain, \eg \cite{langley1992multiple}\cite{langley1992transparent}\cite{darrell1993nulling}.

The flow estimation problem for two-layer images in this paper should not be confused with those works concerning ``motion-layer segmentation", albeit the two do share some similarity and the boundary between them can sometimes be fuzzy. For example, Wang and Adelson~\cite{wang1994representing} proposed to segment the image layers based on a pre-computed optical flow field. Irani \etal \cite{irani1994computing} used temporal integration to track occluding or transparent moving objects with parametric motion. Black and others~\cite{black1996robust}\cite{ju1996skin}\cite{sun2010layered}\cite{wulff2014modeling} proposed a number of algorithms for multiple parametric motion estimation and segmentation.
Yang and Li~\cite{yang2015dense} fit a flow filed with piecewise parametric models.
Weiss~\cite{weiss1997smoothness} presented a nonparametric motion estimation and segmentation method to handle generic smooth motions, thus this method is more related to ours. However, the method of Weiss and most other aforementioned methods primarily focused on
image and motion segmentation,
while we decompose the whole image into two composite brightness layers, and compute one generic flow field on each layer.

The proposed method involves solving two tasks simultaneously:  optical flow field estimation, and reflection/transparent layer separation.  For the second task, many researches have been published previously.
For example,
Levin \etal~\cite{levin2002learning}\cite{levin2007user} proposed methods for separating an image into two transparent layers using local statistics priors of natural images.
Single image solutions are also investigated in \cite{li2014single} and \cite{yeung_CVPR08}.
To utilize multiple frames, layer separation methods have been proposed based on aligning the frames with one layer~\cite{Wexler_ECCV}\cite{Li_brown_ICCV13}\cite{guo2014robust} or multiple layers~\cite{szeliski2000layer}\cite{gai2012blind}.
Sarel and Irani~\cite{sarel2004separating} presented an information theory based approach for separating transparent layers by minimizing the correlation between the layers.  Chen \etal~\cite{Chen_ICCV09} gave a gradient domain approach for moving layer separation which is also based on information theory. Schechner \etal~\cite{schechner2000separation} developed a method for layer separation using image focus as a cue.
By using independent component analysis, Farid and Adelson~\cite{farid1999separating} proposed a layer separation method which works on multiple observations under different mixing weights.
Techniques for image layer separation were also developed in the field of intrinsic image/video extraction~\cite{Tappen_intrinsic}\cite{Weiss_intrinsic}\cite{ye2014intrinsic}.

In the context of stereo matching with transparency, Szeliski and Golland~\cite{szeliski1998stereo} simultaneously recovered disparities, true colors, and opacity of visible surface elements. Tsin \etal~\cite{tsin2006stereo} estimated both depth and colors of the component layers.
Li \etal\cite{li2015simultaneous} proposed a simultaneous video defogging and stereo matching algorithm.

The recent work of Xue \etal~\cite{xue2015computational} has a very similar formulation compared to ours. However, the goal and motivation of obstruction-free photography from a video sequence in \cite{xue2015computational} are different from ours. 
The underlying assumptions on the flow fields, the employed flow solvers and the initialization techniques are
dissimilar.

\section{Problem Setup}
\label{sec:problem}
For ease of presentation, in formulating the problem (Sec.~\ref{sec:problem} and Sec.~\ref{sec:priors}) and presenting the optimization (Sec.~\ref{sec:minimization}),  we will focus on the dynamic foreground case (\ie double-layer flow estimation). The static foreground case (\ie single-layer flow estimation) is simpler and can be derived accordingly. Note that, the static foreground case, though relatively simpler, is also of interest and very challenging.

\subsection{Linear Additive Imaging Model}
In previous discussion, we simply used $\m I=\m L_1\oplus\m L_2$ to denote the layer superposition operation, but did not give its exact form.
To make the idea of this paper more concrete, we opt for the linear additive model $+$ as a concrete example for $\oplus$, \ie, $\m I=\m L_1 + \m L_2$.

The linear additive model itself, while simple, has been used successfully in the past in solving many vision problems involving transparency and reflection (\eg, in shadow removal \cite{yeung_CVPR08}, image matting \cite{szeliski1998stereo} and reflection separation \cite{li2014single}). Moreover, by applying logarithm operation, a multiplicative superposition model can also be converted to an additive one.

Taking two frames of observations, $\m I$ and $\m I'$, at two consecutive time steps $t$ and $t+1$, we have
\begin{eqnarray}
\m I(\m X) \!\!&=&\!\! \m L_1(\m X) +\m L_2 (\m X),\label{eq:linear_additive_model1}\\
\m I'(\m X) \!\!&=&\!\! \m L_1'(\m X) + \m L_2'(\m X), \label{eq:linear_additive_model2}
\end{eqnarray}
where $\m X$ is a matrix indexing all pixel coordinates.
\subsection{Double Layer Brightness Constancy}
In the presence of transparencies or reflections, it is important to note that the conventional BCC condition cannot be applied directly to the observed images.  Below, we will derive a generalized BCC condition which is applicable to the double-layer case.

The basic assumption that we will base our method on is: any component layer of the observed image must satisfy the brightness constancy condition individually. This is a realistic and mild assumption which is applicable to a wide range of transparency and reflection phenomena encountered in natural images. Cases that violate this basic assumption are deemed beyond the scope of this current paper.

Suppose, during two small time steps, layer $\m L_1$ changed to $\m L_1'$ according to a motion field of $\m U$, and layer $\m L_2$ changed to $\m L_2'$ according to a different motion field $\m V$. Based on the assumption that the brightness of the objects in each individual layer is constant, we have
\begin{eqnarray}
\m L_1(\m X) \!\!&=&\!\! \m L_1'(\m X + \m U), \label{eq:double-layer-BCC-1}\\
\m L_2(\m X) \!\!&=&\!\! \m L_2'(\m X + \m V). \label{eq:double-layer-BCC-2}\end{eqnarray}
Together with the imaging model in \eqref{eq:linear_additive_model1} and \eqref{eq:linear_additive_model2}, we call the above constraints the \emph{generalized double-layer BCC condition} for an input double-layer image pair $(\m I, \m I')$.

\subsection{The Double Layer Optical Flow Problem}
Given the above linear additive imaging model as well as the generalized BCC conditions, we aim to recover both $\m L_1$, $\m L_1'$, $\m L_2$, $\m L_2'$ and $\m U$, $\m V$.

To make this severely ill-posed problem trackable, we adopt the energy minimization framework, and base it on the generalized BCC conditions as well as priors for optical flows and image layers. The energy function reads as
\begin{equation}
E  =  E_{B}+\lambda_L E_{L} +\lambda_F E_{F},\label{eq:energy_function}
\end{equation}
where $E_B$ corresponds to the double-layer BCC condition, $E_L$ and $E_F$ are the regularization terms (or prior terms) for the latent image layers, and the unknown optical flow fields, respectively. The $\lambda$s are trade-off parameters.

In energy \eqref{eq:energy_function}, we use $E_B=E_B(\m L_1,\m L_1',\m L_2,\m L_2',\m U,\m V)$ to represent the BCC condition in the following way\footnote{For brevity, hereafter we use a short-hand notation for functions defined on all pixel coordinates $\m X$: a function $f(\m X)$ should be understood as $\sum_{\m x \in \m X}f(\m x)$, unless otherwise specified.}:
\begin{align}
E_B=&\| \m L_1(\m X)-\m L_1'(\m X +\m U) \|+ \|\m L_2(\m X)- \m L_2'(\m X +\m V) \|.
\end{align}
We use $\|\cdot\|$ to denote the $\ell_1$-norm in this paper unless otherwise specified. We choose to use $\ell_1$-norm as the cost function mainly for its robustness~\cite{brox2004high,zach2007duality} and its convenience in optimization. The two regularization terms $E_L$ and $E_F$ will be detailed in the following section.
\section{Regularization}
\label{sec:priors}
Using prior information as regularization is a common practice for solving ill-posed problems.  In this paper, the task is to separate the input frames into latent layers, and to recover the associated flow fields.

Priors are generally task-dependent. By enforcing different priors to latent layers and to optical flow fields, the algorithm can be adapted to solving different tasks.  For example, if one knows the two latent layers are images of natural scenes, then the layers can be assumed to have sparse gradients (\ie, satisfying the well-known natural image priors).  Moreover, for general optical flow fields, one can assume they are piecewise constant or piecewise smooth.

\subsection{Natural Image Prior: Sparse Gradient}
The research in natural image statistics shows that images of typical real-world scenes obey sparse spatial gradient distributions~\cite{Tappen_intrinsic,levin2007user}.  The distribution of a natural image $\m L$ can often be modeled as a generalized Laplace distribution (\emph{a.k.a.}, generalized Gaussian distribution), \ie,
\begin{equation}
P(\m L)= \prod_{\m x\in\m X}\exp(-|\partial_x\m L(\m x)|^p-|\partial_y\m L(\m x)|^p),\label{eq:sparsegrad}
\end{equation}where the power $p$ is a parameter usually within $[0.0,1.0]$. A convenient choice is $p=1$, with which the energy is reduced to the $\ell_1$-norm of image spatial gradients. For ease exposition, we will let $p=1$ in this paper, though bear in mind that using other values of $p$ is possible and may be advantageous in particular applications.

Taking the negative logarithm, the prior in (\ref{eq:sparsegrad}) can be represented in the energy minimization form, \ie
\begin{equation}
\|\nabla\m L(\m X)\|\rightarrow \min,
\end{equation}
where $\nabla=(\partial_x,\partial_y)^\top.$ Therefore, the latent layer regularization term $E_L=E_L(\m L_1,\m L_1',\m L_2,\m L_2')$ can be written as
\begin{align}
E_L\!=\!\|\nabla \m L_1(\m X)\|\!+\!\|\nabla\m L_1'(\m X)\|\!+\!\|\nabla\m L_2(\m X)\|\!+\!\|\nabla\m L_2'(\m X)\|.\label{eq:sprasegradient}
\end{align}

\subsection{Optical Flow Priors: Spatial Smoothness}
Early methods for solving multi-layer optical flow problem often made  restrictive assumption about the unknown flow fields. For example, \cite{bergen1992three} proposed a three-frame algorithm for recovering two component motion fields by assuming that the motion fields are constant over time,  and \cite {shizawa1990simultaneous} was built upon a local constant motion assumption to derive its basic equation.  In this paper, these restrictions are removed and the proposed method can handle more general and more complex motion fields.

We use a general assumption on flow field, namely, the optical flows are generally piecewise constant or piecewise smooth.  To capture this prior, we adopt the total variation (TV) model~\cite{zach2007duality} or total generalized variation (TGV) model~\cite{bredies2010total}. Specifically, a flow field $\m U$ will be regularized by the following energy:
\begin{eqnarray}
\|\m U\|_{{\text{TGV}}^k} \rightarrow \min,
\end{eqnarray}
where $\|\m U\|_{{\text{TGV}}^k}\doteq{\textit{TGV}}^k(\m U_x)+{\textit{TGV}}^k(\m U_y)$, and \mbox{${\textit{TGV}}^k(~\!\!\cdot\!\!~)$} denotes the $k$-th order TGV measure for horizontal and vertical flow components ${\m U}_x$ and ${\m U}_y$.

In general, the $k$-th order TGV favors solutions that are piecewise composed of $(k\!-\!1)$-th order polynomials: with $k=1$, TGV$^1$ reduces to the TV model which favors piecewise constant fields; with $k=2$, TGV$^2$ favors piecewise affine fields. We will only consider TV and TGV$^2$ in this paper, and the resultant prior regularization term $E_F=E_F(\m U,\m V)$ for the flow fields can be written as
\begin{align}
E_F(\m U,\m V) = \|\m U\|_{{\text{TGV}}^k} +\|\m V\|_{{\text{TGV}}^k}.
\end{align}where $k=1$ (\ie TV) or $2$.
\section{Energy Minimization}
\label{sec:minimization}
\subsection{The Overall Objective Function}
By stacking all the constraints over both latent layers and flow fields, we reach an energy minimization problem as
\begin{align}
&\!\!\!\!\!\min E({\m L_1,\m L_1',\m L_2,\m L_2',\m U,\m V})= E_B+\lambda_L E_L +\lambda_F E_F & \nonumber \\
& ~~~~=\!\left(\| \m L_1(\m X)\!-\!\m L_1'(\m X\!+\!\m U) \|+ \|\m L_2(\m X)\!-\!\m L_2'(\m X\!+\!\m V) \|\right) & \nonumber \\
& ~~~~~~~~+ \lambda_L \left(\|\nabla\m L_1\|\!+\!\|\nabla\m L_1'\|\!+\!\|\nabla\m L_2\|\!+\!\|\nabla\m L_2'\|\right) & \nonumber\\
& ~~~~~~~~+ \lambda_F \left(\|\m U\|_{{\text{TGV}}^k} +\|\m V\|_{{\text{TGV}}^k}\right), & \label{eq:objfunction}\\
&\!\!\!\!\text{subject to} \nonumber & \\
& ~~~~~~~~~~~~~~~~~~\m I = \m L_1 +\m L_2 ,~~\m I' = \m L_1' + \m L_2',& \label{eq:ADD} \\
& ~~~~~~~\m 0\le\m L_2\le\min(\m I, c),~~\m 0\le\m L_2'\le \min(\m I', c). &\label{eq:bound}
\end{align}where the $\m X$'s in the gradient terms of \eqref{eq:sprasegradient} are omitted for brevity.

Note that, to distinguish background and foreground layers, we introduce in \eqref{eq:bound} the element-wise bound constraints on the layers. We assume the foreground layer containing transparency or reflection has weaker signal, and use a small constant scalar $c$ (\eg $c=0.25$ for brightness values in the range of [0,1]) as its brightness upper bound. This can be understood as an additional bound prior for layer separation. Also note that, putting aside \eqref{eq:bound}, there is a global shift ambiguity for the layer values: adding an arbitrary scalar $s\in\mathbb{R}$ to $\m L_1,\m L_1'$ then $-s$ to $\m L_2,\m L_2'$ dose not change the energy in (\ref{eq:objfunction}), nor dose it affect (\ref{eq:ADD}). This is because all the terms in (\ref{eq:objfunction}) depend on value difference rather than absolute value. Nevertheless, both the lower and upper bounds in \eqref{eq:bound} help constrain the absolute values.
\subsection{Alternated Minimization}
To solve the above energy minimization problem, we first substitute the additive model constraints in \eqref{eq:ADD} as hard constraints to eliminate $\m L_1$ and $\m L_1'$ in (\ref{eq:objfunction}). Consequently, the energy function is now defined only on latent layers $\m L_2,\m L_2'$ and optical flows ${\m U, \m V}$.

Then, examining the energy form in \eqref{eq:objfunction}, we notice that: \emph{i}) the prior terms for optical flow field, \ie $E_F$, is independent of the prior term for latent layers $E_L$; and \emph{ii}) the BCC energy term $E_B$ is the only term that links the flow estimation with latent layer separation. Based on these observations, we solve the minimization problem via block coordinate descent in an alternating fashion.

Specifically, starting from a proper initialization, our algorithm alternately solves the following two sub-problems:
\begin{itemize}
\item {\bf(Layer Separation):} Given current flow field estimates $\{\m U,\m V\}$, solve for image layers \{$\m L_2, \m L_2'$\} via the following minimization:
\vspace{-1pt}
\begin{eqnarray}
\min_{\m L_2,\m L_2'}\left( E_B(\m L_2, \m L_2' ) + \lambda_L E_L(\m L_2, \m L_2') \right).
\end{eqnarray}
\item {\bf(Flow Computation):} Given current image layers $\{\m L_2,\m L_2'\}$, estimate $\{\m U,\m V\}~$ by solving the following two-layer optical flow problem:
\vspace{-1pt}
\begin{eqnarray}
\min_{\m U, \m V} \left( E_{B}(\m U, \m V) +\lambda_F E_{F}(\m U, \m V) \right).
\end{eqnarray}
\end{itemize}

\vspace{-1pt}
More details are given below.
\vspace{-4pt}
\subsubsection{Update the image layers}
Given current optical flow estimates $\m U$ and $\m V$, the latent image layers $\m L_2, \m L_2^{'}$ can be updated by solving the following optimization problem:
\begin{align}
&\!\!\!\!\!\!\min_{\m L_2,\m L_2'}\!\! \|(\m I \!-\! \m L_2)(\m X)\!-\! (\m I' \!\!-\!\! \m L_2')(\m X \!\!+\!\!\m U) \|\!\!+\!\!\| \m L_2(\m X)\! -\!\m L_2'(\m X\!\!+\!\!\m V) \| & \nonumber \\
& ~~~~+\!\lambda_L\! \left( \|\nabla(\m I \!-\! \m L_2)\|\!+\!\|\nabla(\m I' \!-\! \m L_2')\|\!+\!\|\nabla\m L_2\| \!+\!\|\nabla\m L_2'\|\right) & \nonumber \\
&\!\!\!\!\mbox{subject to~~} \m 0\!\le\!\m L_2\!\le\!\min(\m I, c),~\m 0\!\le\!\m L_2'\!\le\!\min(\m I', c),&
\end{align}
This is a convex optimization problem defined on $\m L_2$ and $\m L_2'$, and the cost function can be arranged into
\begin{eqnarray}
&\displaystyle\min_{\mathbf{l}} \| \m A \cdot \mathbf{l}- \mathbf{b}\|,\nonumber\\
&\mbox{subject to~~} lb_i \leq l_i \leq ub_i, \forall i
\end{eqnarray}where $\m A$ and $\mathbf{b}$ encode all the $\ell_1$ constraints on latent layers, which are extremely sparse (only a few elements in each row are non-zero). $\mathbf{l}$ is a column vector containing elements in $\m L_2$ and $\m L_2'$. $lb_i$ and $ub_i$ are constant bounds from (\ref{eq:bound}). The constraints are linear function of the latent layers $\m L_2$ and $\m L_2'$, thus this problem can be solved as a linear programming using off-the-self solvers.

Nevertheless, to utilize the sparse structure in the problem and speed up the implementation, we solve the problem by using a tailored version of Iteratively Reweighted Least Squares (IRLS)~\cite{chartrand2008iteratively}.  With IRLS, one can also adapt the formulation to different priors readily, \eg, replacing $\ell_1$-norm with $\ell_p$-norm ($0\!<\!p\!<\!1$). Details of our IRLS variant can be found in the \emph{Supplementary Material}.

\paragraph{Use of color images.} The above formulations can be easily extended to color RGB images.  With color images, the double-layer BCC term $E_B$ and layer regularization term $E_L$ will be evaluated at R-G-B channels separately.  The flow fields $\m U$ and $\m V$ are shared by all three channels.

\subsubsection{Update the flow fields $\m U$ and $\m V$}
Given current layer estimates $\m L_2, \m L_2'$, and $\m L_1 = \m I - \m L_2$, $\m L_1' = \m I'- \m L_2'$, the next step is to update the associated two flow fields $\m U$ and $\m V$. This is done by solving the following optimization problem:
\begin{align}
& \min_{\m U, \m V} \| \m L_1(\m X) -\m L_1'(\m X +\m U) \|+ \|\m L_2(\m X)-\m L_2'(\m X +\m V) \| & \nonumber \\  &~~~~+\lambda_F \left(\|\m U\|_{{\text{TGV}}^k} +\|\m V\|_{{\text{TGV}}^k}\right).&
\end{align}

The computations for these two flow fields are in fact separable.  This can be easily seen from the above optimization, as the cost function can be expressed as the sum of two terms, each of which can be solved in isolation, \ie, given $\{\m L_1,\m L_1',\m L_2,\m L_2'\},$ solve
\begin{align}
& ~~~~~\min_{\m U} \| \m L_1(\m X) -\m L_1'(\m X +\m U) \| +\lambda_F \|\m U\|_{{\text{TGV}}^k}, & \\
& ~~~~~\min_{\m V} \|\m L_2(\m X)- \m L_2'(\m X +\m V) \| +\lambda_F \|\m V\|_{{\text{TGV}}^k}. &
\end{align}
To solve the above optical flow problems, we use quadratic relaxation and introduce an auxiliary flow field to decouple the BCC term and regularization term, similar to \cite{zach2007duality,steinbrucker2009large}.  To solve the resulting TV-$L_2$ (\emph{a.k.a.} the ROF model) and TGV$^2$-$L_2$ problem, we apply the primal-dual method of \cite{chambolle2011first} which is GPU-friendly. Details of our algorithm and implementation can be found in the \emph{Supplementary Material}.

\section{Experimental Results}
\begin{figure}[!tp]
\begin{center}
  \includegraphics[width=0.48\textwidth]{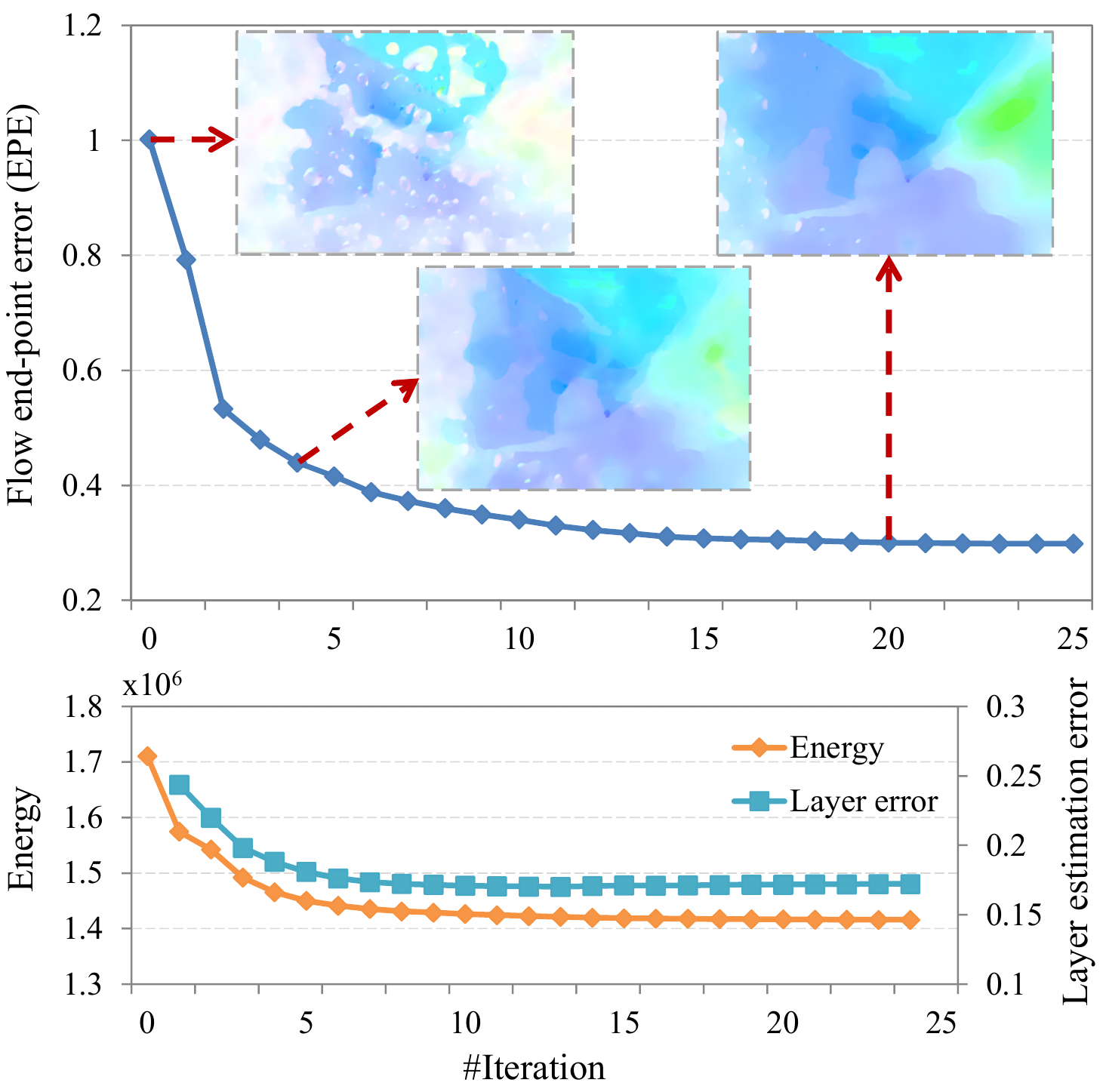}
\end{center}
\vspace{-13pt}
\caption{Convergence of the proposed method. Top: optical flow estimation error (EPE) \emph{w.r.t.} iterations. Bottom: energy and layer estimation errors \emph{w.r.t.} iterations. The layer error is evaluated as $1-NCC(\text{GT}~\m L_2, \text{estimated}~\m L_2$).\label{fig:convergence}}
\vspace{2pt}
\end{figure}

In this section, we validate the proposed model and framework, and evaluate the performance of our method. We report the experimental results on both synthetic data and real images (\eg Middlebury~\cite{middlebury} and Sintel~\cite{butler2012naturalistic} flow datasets, and the reflection dataset in \cite{Li_brown_ICCV13}).


\paragraph{Initialization.} Being an alternated method, the proposed algorithm requires an initialization to start the alternation. One can start from either an initial optical flow estimation or from an initial layer separation. The latter one is used in our experiments, and the initialization details will be given later in the experiments.
\paragraph{Parameters.} In the following experiments, the weights of the priors, \ie $\lambda_L$, $\lambda_F$, are roughly tuned according to the results. Both TV and TGV$^2$ flow regularizers worked well, consistently improving the accuracy upon initialization. Due to space limitation, in the following we report the results using TV (\ie $k=1$). The results using TGV$^2$ (\ie $k=2$) can be found in the \emph{Supplementary Material}.

\subsection{Static Foreground Cases}
\begin{figure*}[!htp]
  \begin{center}
  \begin{subfigure}{0.162\linewidth}
  \includegraphics[width=1\textwidth]{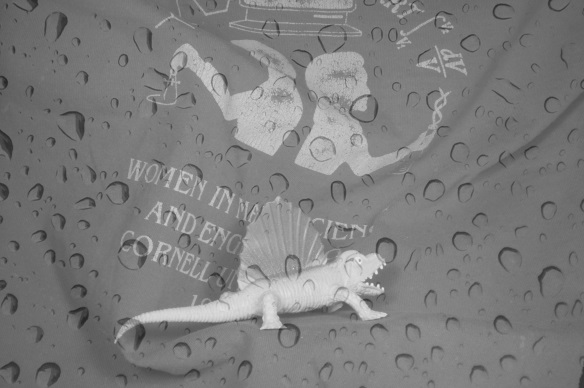}
  \vspace{-17pt}\caption{\footnotesize Input $\m I$}
  \end{subfigure}
  \begin{subfigure}{0.162\linewidth}
  \includegraphics[width=1\textwidth]{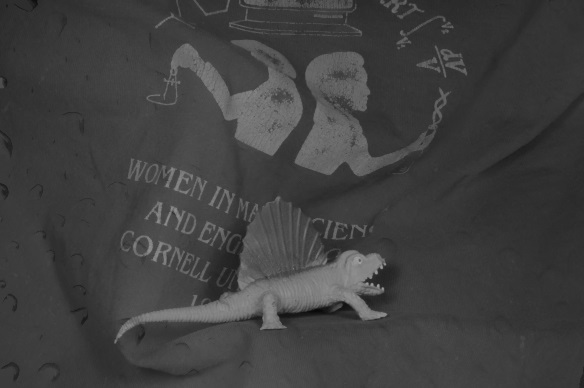}
  \vspace{-17pt}\caption{\footnotesize Output $\m L_1$}
  \end{subfigure}
  \begin{subfigure}{0.162\linewidth}
  \includegraphics[width=1\textwidth]{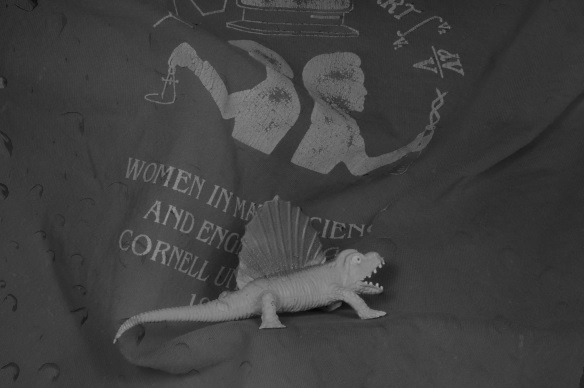}
  \vspace{-17pt}\caption{\footnotesize Output $\m L_1'$}
  \end{subfigure}
  \begin{subfigure}{0.162\linewidth}
  \includegraphics[width=1\textwidth]{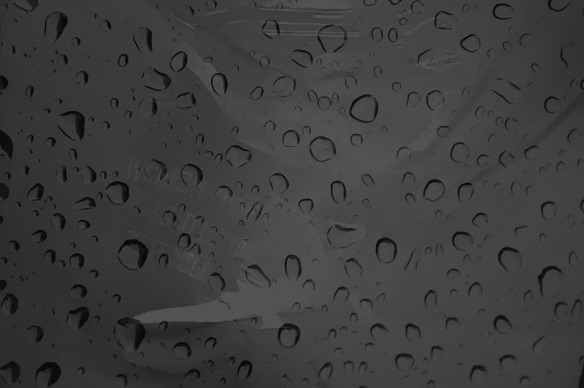}
  \vspace{-17pt}\caption{\footnotesize Output $\m L_2$}
  \end{subfigure}
  \begin{subfigure}{0.162\linewidth}
  \includegraphics[width=1\textwidth]{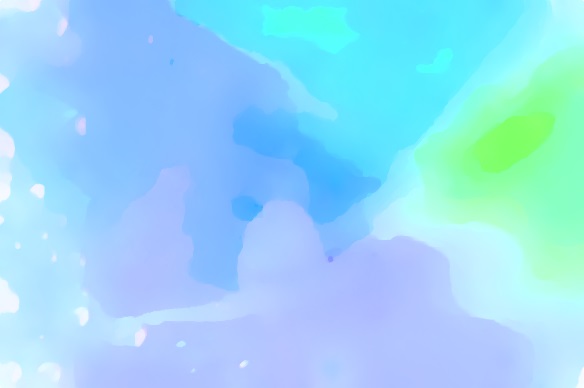}
  \vspace{-17pt}\caption{\footnotesize Output $\m U$ (epe 0.29)}
  \end{subfigure}
  \begin{subfigure}{0.162\linewidth}
  \includegraphics[width=1\textwidth]{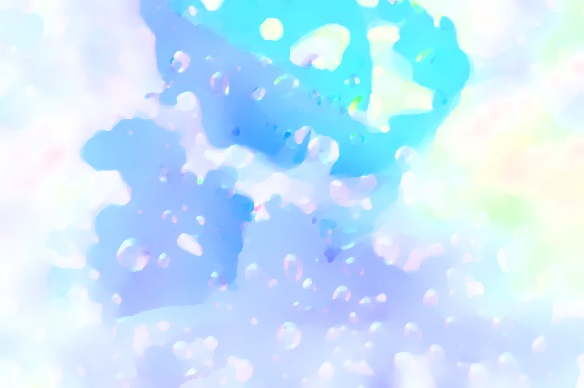}
  \vspace{-17pt}\caption{\footnotesize Naive $\m U$ (epe 1.01)}
  \end{subfigure}\\
  \vspace{2pt}
  \begin{subfigure}{0.162\linewidth}
  \includegraphics[width=1\textwidth]{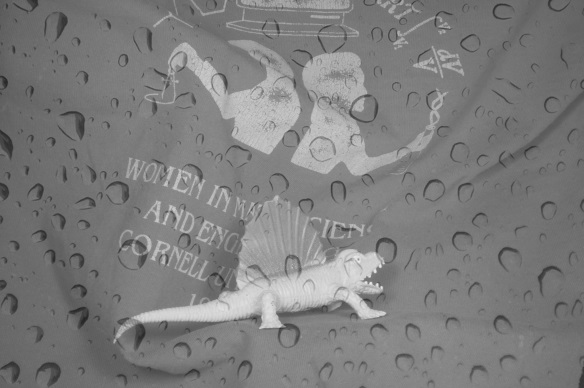}
  \vspace{-17pt}\caption{\footnotesize Input $\m I'$}
  \end{subfigure}
  \begin{subfigure}{0.162\linewidth}
  \includegraphics[width=1\textwidth]{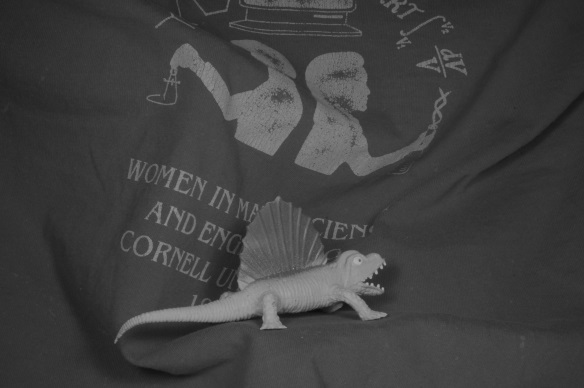}
  \vspace{-17pt}\caption{\footnotesize GT $\m L_1$}
  \end{subfigure}
  \begin{subfigure}{0.162\linewidth}
  \includegraphics[width=1\textwidth]{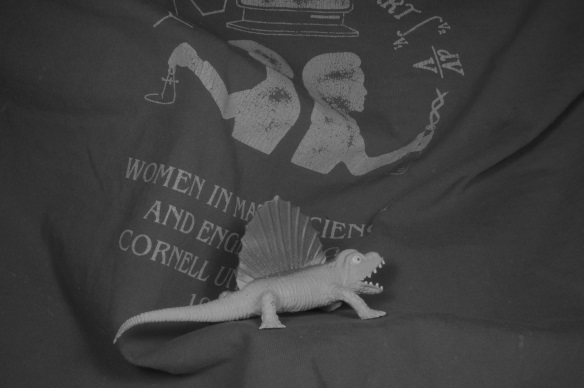}
  \vspace{-17pt}\caption{\footnotesize GT $\m L_1'$}
  \end{subfigure}
  \begin{subfigure}{0.162\linewidth}
  \includegraphics[width=1\textwidth]{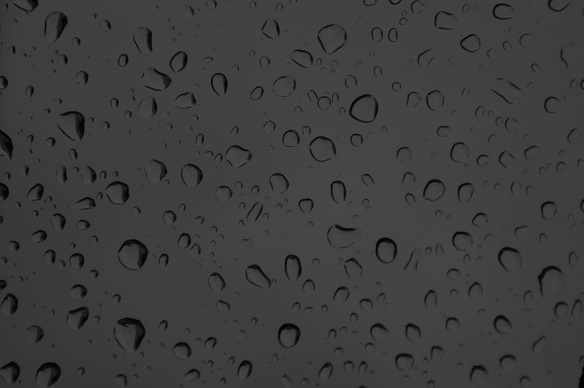}
  \vspace{-17pt}\caption{\footnotesize GT $\m L_2$}
  \end{subfigure}
  \begin{subfigure}{0.162\linewidth}
  \includegraphics[width=1\textwidth]{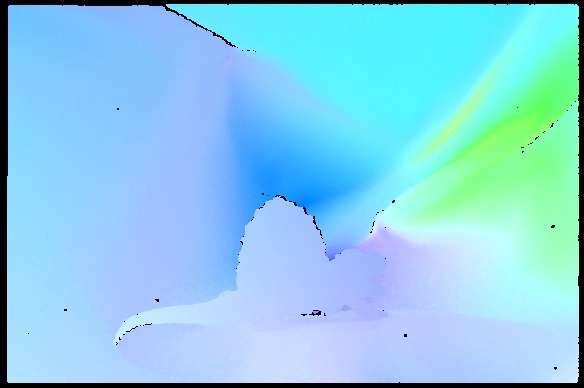}
  \vspace{-17pt}\caption{\footnotesize GT $\m U$}
  \end{subfigure}
  \begin{subfigure}{0.162\linewidth}
  \includegraphics[width=1\textwidth]{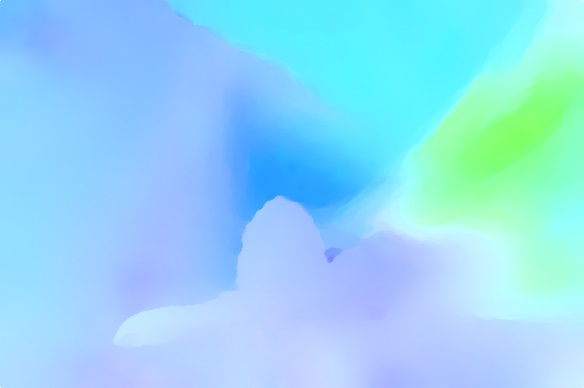}
  \vspace{-17pt}\caption{\footnotesize Oracle $\m U$ (epe 0.16)}
  \end{subfigure}
  \end{center}
  \vspace{-15pt}
  \caption{Performance evaluation of the proposed method on a single flow case, where a rain image is superimposed on the Dimetrodon image pair. The estimated flow (e) is significantly better than the initialization (f), a naive optical flow estimate without layer separation. The error evolution curve is shown in Fig.~\ref{fig:convergence}.  Oracle flow (l) is computed with clean background images (\ie with ground-truth layer separations). (\emph{\textbf{Best viewed on screen}})}
  \label{fig:single_gray_cvpr16}
\end{figure*}
\begin{figure*}[!htp]
  \begin{center}
  \begin{subfigure}{0.195\linewidth}
  \includegraphics[width=1\textwidth]{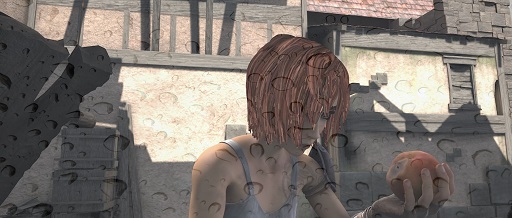}
  \vspace{-17pt}\caption{\footnotesize Input $\m I$}
  \end{subfigure}
  \begin{subfigure}{0.195\linewidth}
  \includegraphics[width=1\textwidth]{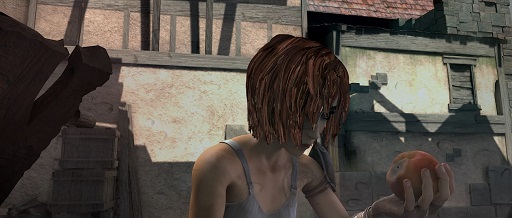}
  \vspace{-17pt}\caption{\footnotesize Output $\m L_1$}
  \end{subfigure}
  \begin{subfigure}{0.195\linewidth}
  \includegraphics[width=1\textwidth]{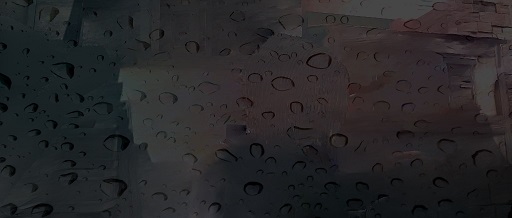}
  \vspace{-17pt}\caption{\footnotesize Output $\m L_2$}
  \end{subfigure}
  \begin{subfigure}{0.195\linewidth}
  \includegraphics[width=1\textwidth]{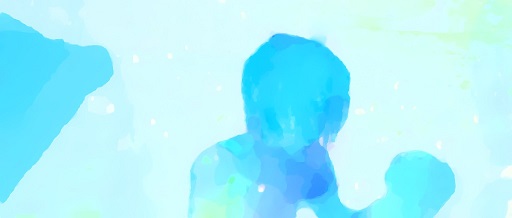}
  \vspace{-17pt}\caption{\footnotesize Output $\m U$ (epe 0.30)}
  \end{subfigure}
  \begin{subfigure}{0.195\linewidth}
  \includegraphics[width=1\textwidth]{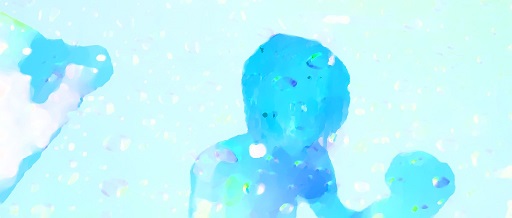}
  \vspace{-17pt}\caption{\footnotesize Naive $\m U$ (epe 0.61)}
  \end{subfigure}\\
  \begin{subfigure}{0.195\linewidth}
  \includegraphics[width=1\textwidth]{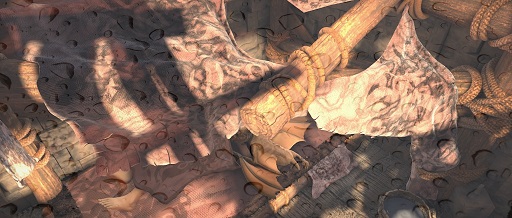}
  \vspace{-17pt}\caption{\footnotesize Input $\m I$}
  \end{subfigure}
  \begin{subfigure}{0.195\linewidth}
  \includegraphics[width=1\textwidth]{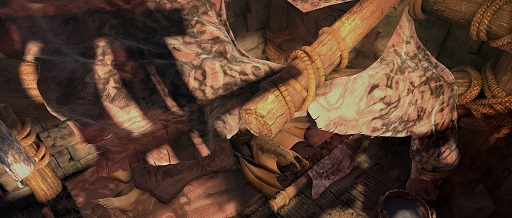}
  \vspace{-17pt}\caption{\footnotesize Output $\m L_1$}
  \end{subfigure}
  \begin{subfigure}{0.195\linewidth}
  \includegraphics[width=1\textwidth]{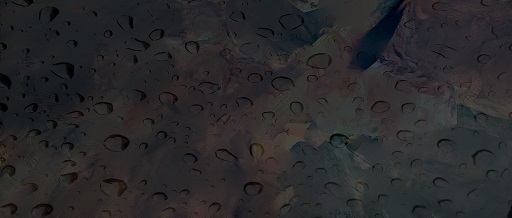}
  \vspace{-17pt}\caption{\footnotesize Output $\m L_2$}
  \end{subfigure}
  \begin{subfigure}{0.195\linewidth}
  \includegraphics[width=1\textwidth]{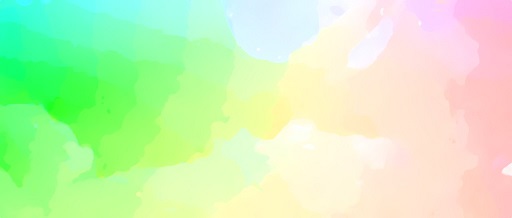}
  \vspace{-17pt}\caption{\footnotesize Output $\m U$ (epe 0.21)}
  \end{subfigure}
  \begin{subfigure}{0.195\linewidth}
  \includegraphics[width=1\textwidth]{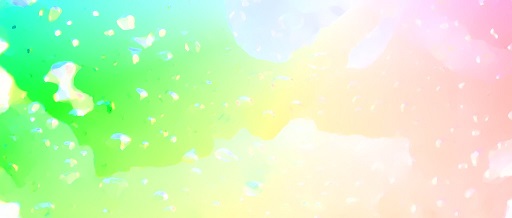}
  \vspace{-17pt}\caption{\footnotesize Naive $\m U$ (epe 0.33)}
  \end{subfigure}
  \end{center}
  \vspace{-15pt}
  \caption{Typical results of our method on single-flow cases, where the rain drop image is superimposed on images from the Sintel dataset. For clarity, we only show here the first frame $\m I$ and its layer separation result. (\emph{\textbf{Best viewed on screen}})}
  \label{fig:sintel}
\end{figure*}

\begin{table}
  \centering
  \caption{Mean flow EPE for three Sintel image sequences superimposed with the static rain image.  Oracle flows are computed with clean background images.}\label{tab:sintelepe}
  \vspace{-5pt}
  \begin{tabular}{|c|c|c|c|}
  \hline
  Sequence & Naive flow & Our flow & Oracle \\
  \hline
  ``alley1" & 0.49 & \textbf{0.35} & 0.22 \\
  \hline
  ``sleeping1" & 0.80 & \textbf{0.33} & 0.12 \\
  \hline
  ``sleeping2" & 0.26 & \textbf{0.21} & 0.07 \\
  \hline
  \end{tabular}
\end{table}

We start from the simpler case where only background layer $\m L_1$ is dynamically changing by an unknown motion field $\m U$, while the foreground layer is static (\ie $\m L_2 \equiv \m L_2'$ and $\m V \equiv \m 0$).
The task is to estimate flow field $\m U$ and component layers $\m L_1,\m L_1',\m L_2$. Again, we would like to emphasize that, even though we call it the ``simpler case'', to jointly estimate an accurate flow field and recover latent layers remains a challenging task. To the best of our knowledge, there was no previous method that recovers both a complex dense flow field under transparency/reflection, and separate the two constituting layers.

In the following tests, a rather conservative strategy is used to initialize the proposed method: we initiate the static foreground image $\m L_2$ to be all zeros. Consequently, in the beginning of the optimization we compute an initial optical flow field naively based on the two input images.

Seeing through rain is a practical situation where measures should be taken to avoid the rain ruining vision systems. In the first test, we first synthesized a scene by superimposing a static rain image over the pair of Dimetrodon in the  Middlebury dataset. Gray images were used. As illustrated in Fig.~\ref{fig:convergence}, within about 25 iterations, the optical flow estimation error has been decreased from about 1.0 pixels to about 0.3 pixels. This demonstrates the advantage of our formulation for robust optical flow estimation. The qualitative results are demonstrated in Fig.~\ref{fig:single_gray_cvpr16}.

Additionally, we overlay the rain image with three color image sequences from the Sintel dataset. We evenly sampled 10 images from the ``alley 1", ``sleeping 1" , and ``sleeping 2" sequences respectively, and Table~\ref{tab:sintelepe} shows that the proposed method has clearly reduced the mean EPE of initial flows. Two typical results are shown in Fig.~\ref{fig:sintel}.

\begin{figure*}[!htp]
  \vspace{0pt}
  \begin{center}
  \begin{subfigure}{0.162\linewidth}
  \includegraphics[width=1\textwidth]{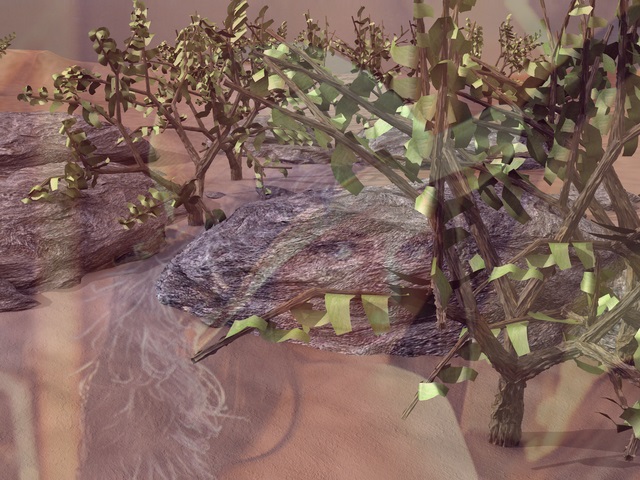}
  \vspace{-17pt}\caption{\footnotesize Input $\m I$}
  \end{subfigure}
  \begin{subfigure}{0.162\linewidth}
  \includegraphics[width=1\textwidth]{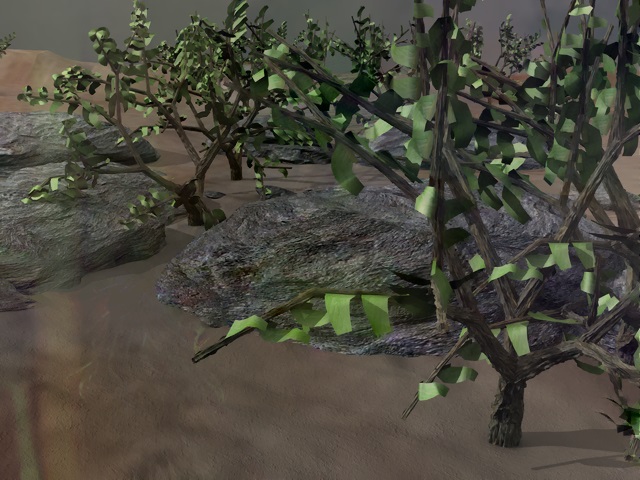}
  \vspace{-17pt}\caption{\footnotesize Output $\m L_1$}
  \end{subfigure}
  \begin{subfigure}{0.162\linewidth}
  \includegraphics[width=1\textwidth]{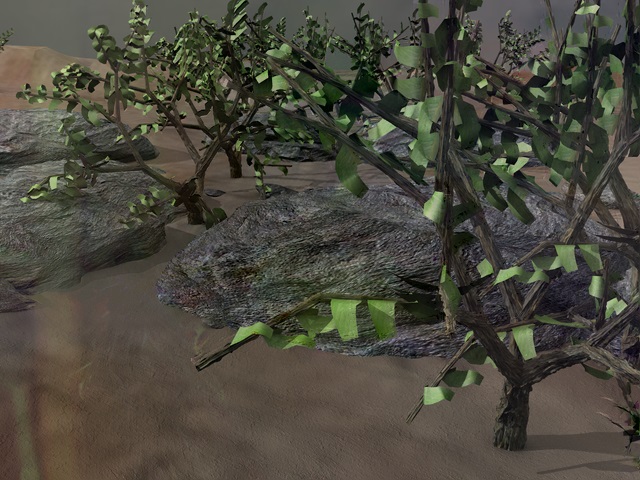}
  \vspace{-17pt}\caption{\footnotesize Output $\m L_1'$}
  \end{subfigure}
  \begin{subfigure}{0.162\linewidth}
  \includegraphics[width=1\textwidth]{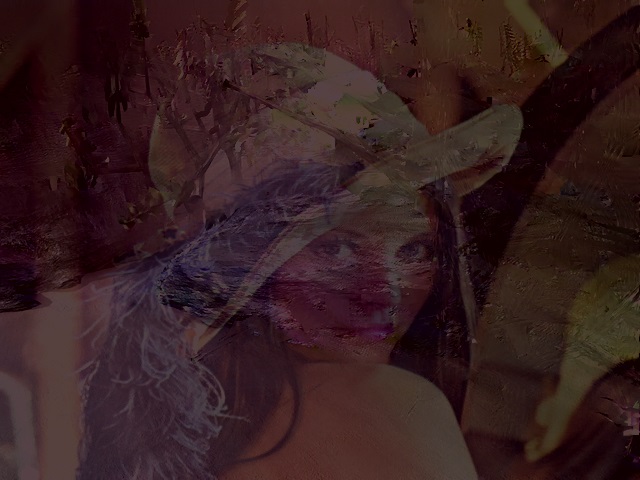}
  \vspace{-17pt}\caption{\footnotesize Output $\m L_2$}
  \end{subfigure}
  \begin{subfigure}{0.162\linewidth}
  \includegraphics[width=1\textwidth]{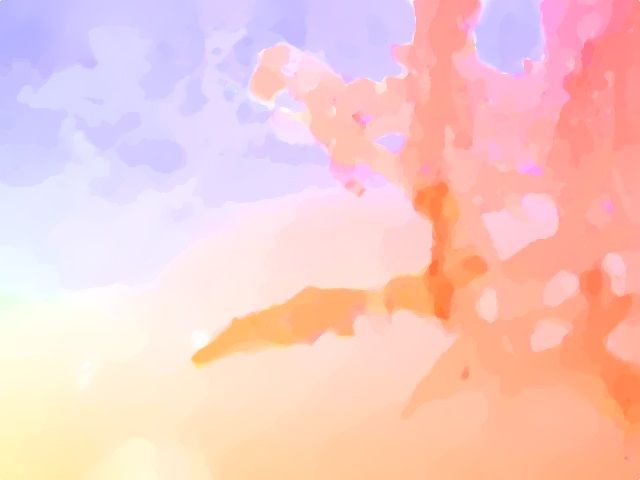}
  \vspace{-17pt}\caption{\footnotesize Output $\m U$ (epe 0.88)}
  \end{subfigure}
  \begin{subfigure}{0.162\linewidth}
  \includegraphics[width=1\textwidth]{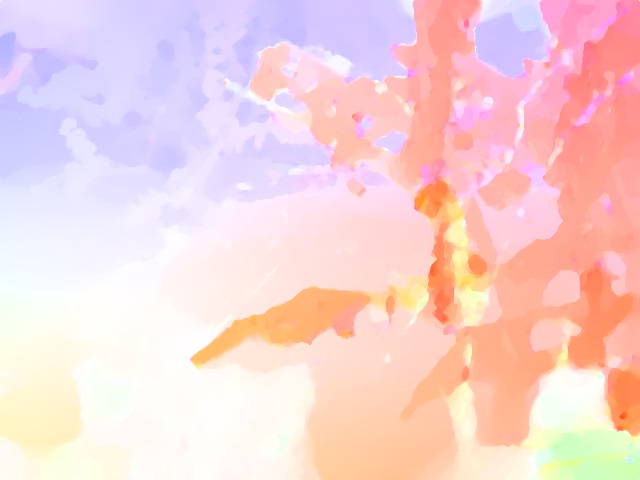}
  \vspace{-17pt}\caption{\footnotesize Naive $\m U$ (epe 1.45)}
  \end{subfigure}\\
  \vspace{2pt}
  \begin{subfigure}{0.162\linewidth}
  \includegraphics[width=1\textwidth]{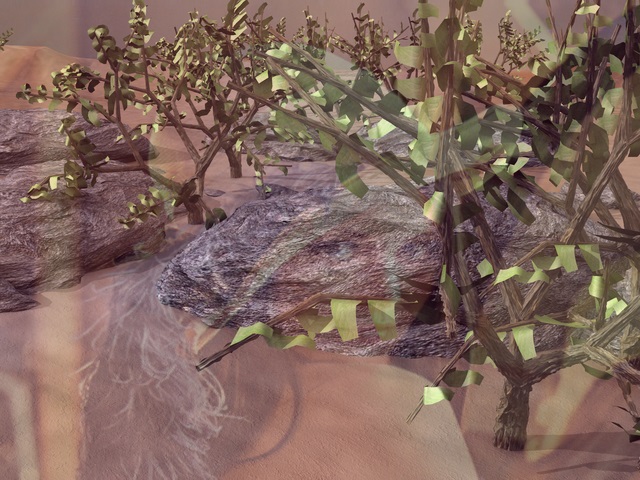}
  \vspace{-17pt}\caption{\footnotesize Input $\m I'$}
  \end{subfigure}
  \begin{subfigure}{0.162\linewidth}
  \includegraphics[width=1\textwidth]{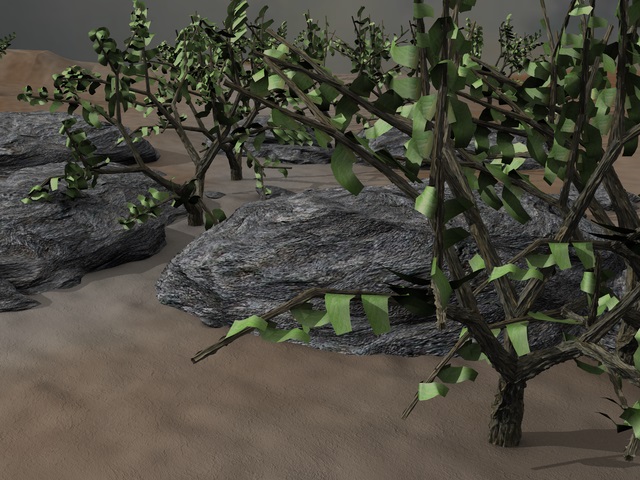}
  \vspace{-17pt}\caption{\footnotesize GT $\m L_1$}
  \end{subfigure}
  \begin{subfigure}{0.162\linewidth}
  \includegraphics[width=1\textwidth]{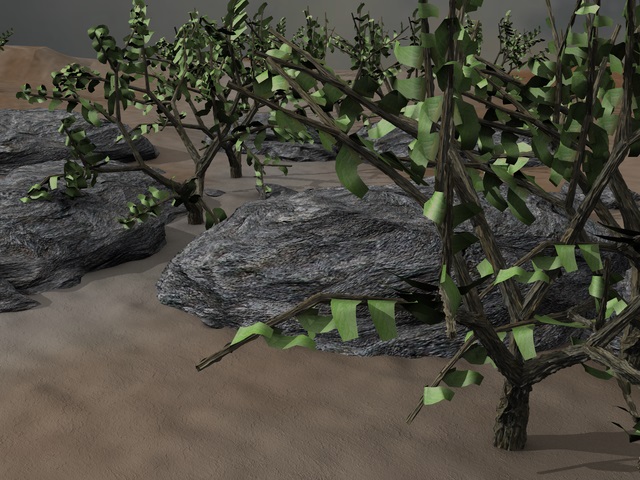}
  \vspace{-17pt}\caption{\footnotesize GT $\m L_1'$}
  \end{subfigure}
  \begin{subfigure}{0.162\linewidth}
  \includegraphics[width=1\textwidth]{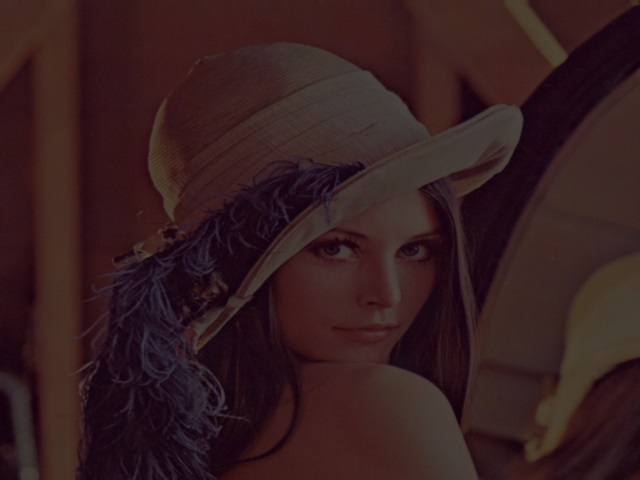}
  \vspace{-17pt}\caption{\footnotesize GT $\m L_2$}
  \end{subfigure}
  \begin{subfigure}{0.162\linewidth}
  \includegraphics[width=1\textwidth]{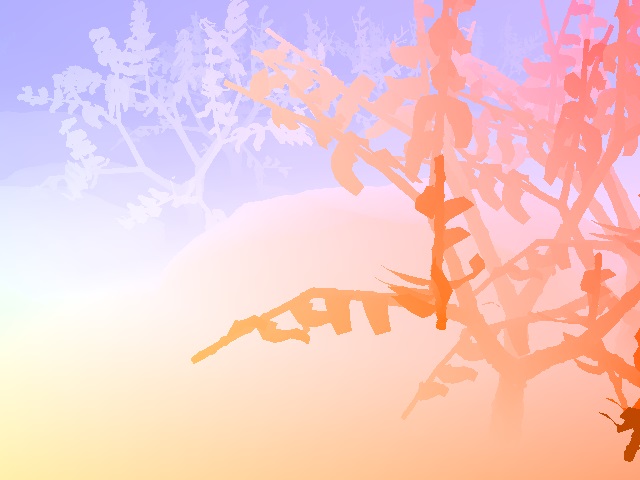}
  \vspace{-17pt}\caption{\footnotesize GT $\m U$}
  \end{subfigure}
  \begin{subfigure}{0.162\linewidth}
  \includegraphics[width=1\textwidth]{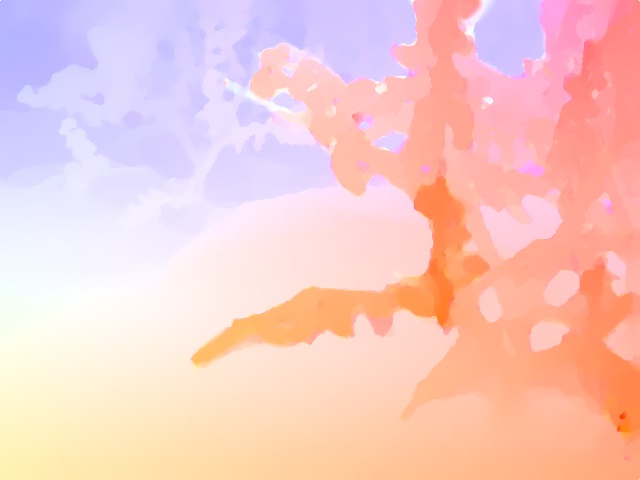}
  \vspace{-17pt}\caption{\footnotesize Oracle $\m U$ (epe 0.67)}
  \end{subfigure}
  \end{center}
  \vspace{-15pt}
  \caption{Performance evaluation of the proposed method on a single flow case, where the Lena image is superimposed on the Grove image pair. The estimated flow (e) is significantly better than the initialization (f), a naive optical flow estimate without layer separation. Oracle flow (l) is computed with ground-truth $\m L_2$. (\emph{\textbf{Best viewed on screen}})}
  \label{fig:single_color_5}
  \vspace{0pt}
\end{figure*}

To further test the performance of our method, we synthesized another pair by superimposing the Lena image with the Grove image in the Middlebury dataset. The results are demonstrated in Fig.~\ref{fig:single_color_5}. Again, we obtained a much better optical flow compared to the initial naive optical flow estimate. As for the layer separation results, the portrait of Lena can be hardly seen in the restored grove images.

\begin{figure}[!tp]
  \begin{center}
  \begin{subfigure}{0.314\linewidth}
  \includegraphics[width=1\textwidth]{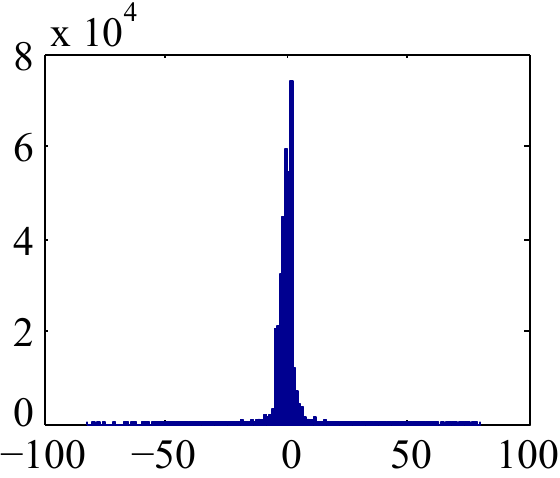}
  \vspace{-17pt}\caption{\footnotesize Dimetrodon}
  \end{subfigure}
  \begin{subfigure}{0.32\linewidth}
  \includegraphics[width=1\textwidth]{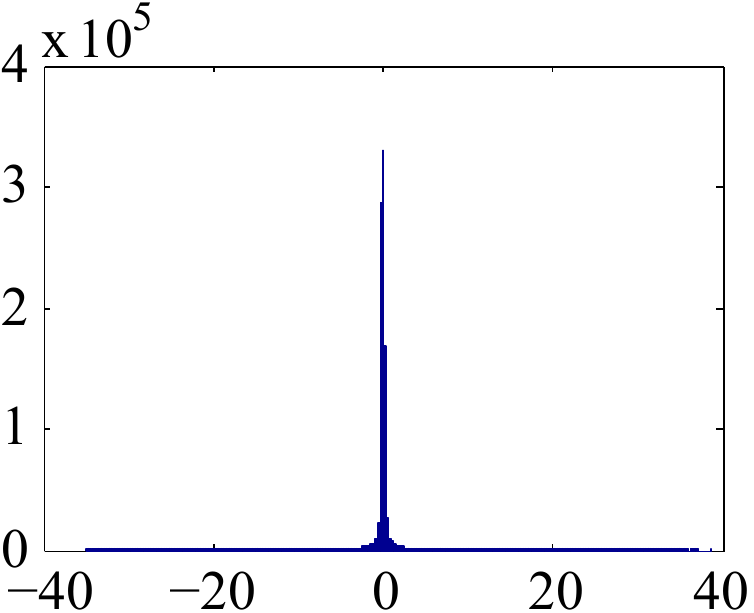}
  \vspace{-17pt}\caption{\footnotesize Rain drop}
  \end{subfigure}\!
  \begin{subfigure}{0.33\linewidth}
  \includegraphics[width=1\textwidth]{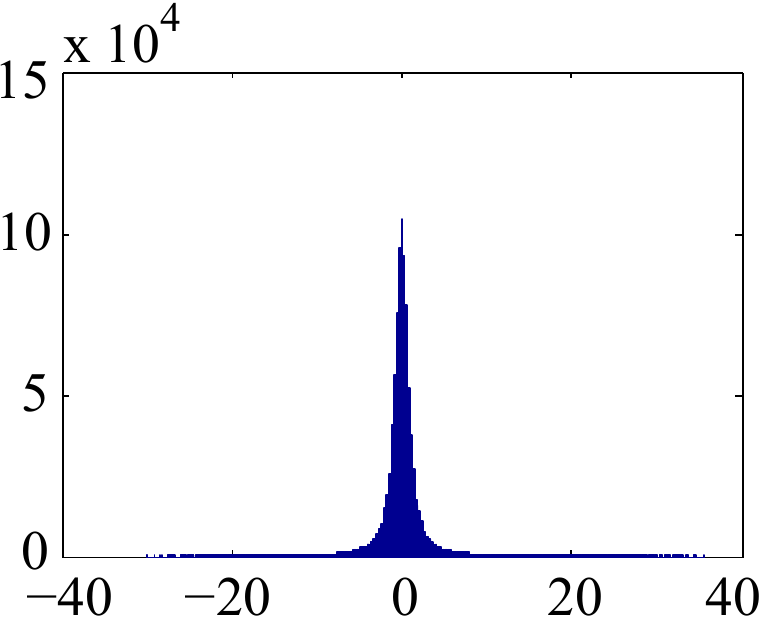}
  \vspace{-17pt}\caption{\footnotesize Lena}
  \end{subfigure}
  \end{center}
  \vspace{-13pt}
  \caption{Gradient statistics of three used images.}
  \label{fig:statistics}
  \vspace{0pt}
\end{figure}

In Fig.~\ref{fig:statistics} we show the image gradient statistics of the three foreground images used in the above experiments. The experimental results have shown that the proposed method works well on these images with the sparse gradient prior. Whenever available, other strong statistical priors can be incorporated into the optimization framework to further improve the performance.

\subsection{Dynamic Foreground Cases}
In this section, we test the proposed method in the dynamic foreground cases, where the task is that given two frames of input images $\m I$ and $\m I'$, recover four component layers $\m L_1, \m L_1', \m L_2, \m L_2'$, and two dense motion fields $\m U$, $\m V$. In the problem of reflection removal, both the background scene and the reflection can be dynamic, which can give rise to such a situation.


We use two pairs of dynamic reflection scenes from \cite{Li_brown_ICCV13} to test the proposed method on the double-layer optical flow problem. In previous single-flow experiments, we initialize the method with foreground layers being all zero. However, this simple strategy did not work for the double-flow case. No reasonably good flow field could be obtained with this strategy for the background or reflection layer, especially for the reflection layer as its signal is weak. Indeed, the fact that the background layer is much more prominent has been took advantage of by some layer separation methods~\cite{Li_brown_ICCV13}\cite{guo2014robust} which align the input images with respect to the background layer.
To obtain proper initialization, we first ran method of \cite{Li_brown_ICCV13} for initial layer separations\footnote{Method of \cite{Li_brown_ICCV13} takes multiple images as input, with one of them being the reference on which the reflection is to be removed. We apply this method on two images, and run it twice with each image as reference.}, then computed initial optical flows on them.

\begin{figure*}[!htp]
\vspace{-5pt}
\begin{center}
\captionsetup[subfigure]{labelformat=empty}
  \begin{subfigure}{0.138\linewidth}
  \includegraphics[width=1\textwidth]{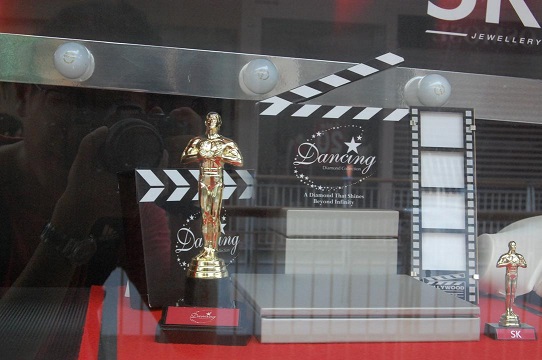}
  \vspace{-18pt}\caption{\footnotesize Input $\m I$}
  \end{subfigure}
  \begin{subfigure}{0.138\linewidth}
  \includegraphics[width=1\textwidth]{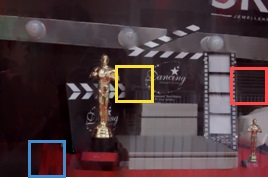}
  \vspace{-18pt}\caption{\footnotesize Initial $\m L_1$}
  \end{subfigure}
  \begin{subfigure}{0.138\linewidth}
  \includegraphics[width=1\textwidth]{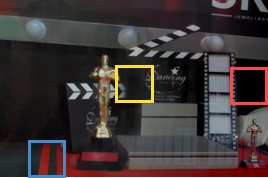}
  \vspace{-18pt}\caption{\footnotesize Final $\m L_1$}
  \end{subfigure}
  \begin{subfigure}{0.138\linewidth}
  \includegraphics[width=1\textwidth]{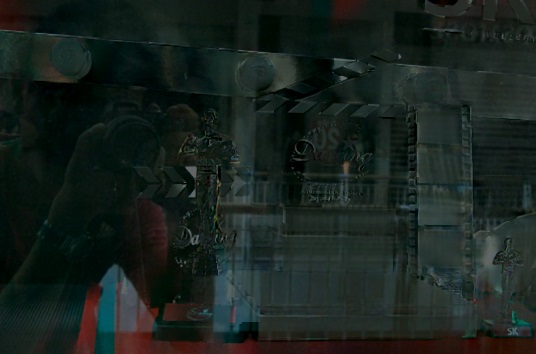}
  \vspace{-18pt}\caption{\footnotesize Initial $\m L_2$}
  \end{subfigure}
  \begin{subfigure}{0.138\linewidth}
  \includegraphics[width=1\textwidth]{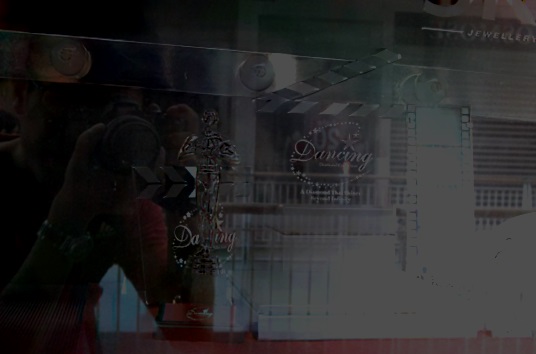}
  \vspace{-18pt}\caption{\footnotesize Final $\m L_2$}
  \end{subfigure}
  \begin{subfigure}{0.138\linewidth}
  \includegraphics[width=1\textwidth]{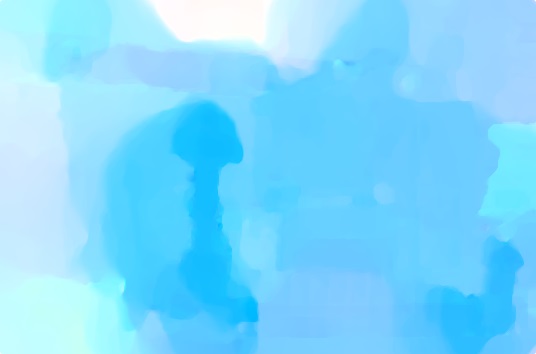}
  \vspace{-18pt}\caption{\footnotesize Initial $\m U$}
  \end{subfigure}
  \begin{subfigure}{0.138\linewidth}
  \includegraphics[width=1\textwidth]{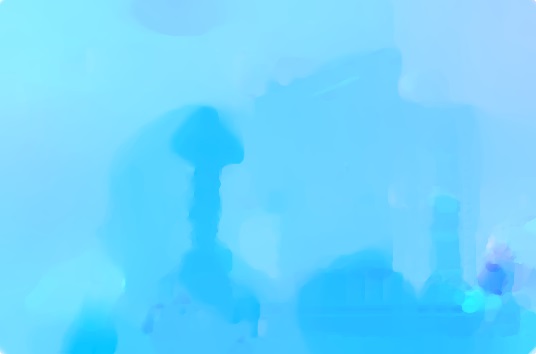}
  \vspace{-18pt}\caption{\footnotesize Final $\m U$}
  \end{subfigure}\\
  \vspace{2pt}
  \begin{subfigure}{0.138\linewidth}
  \includegraphics[width=1\textwidth]{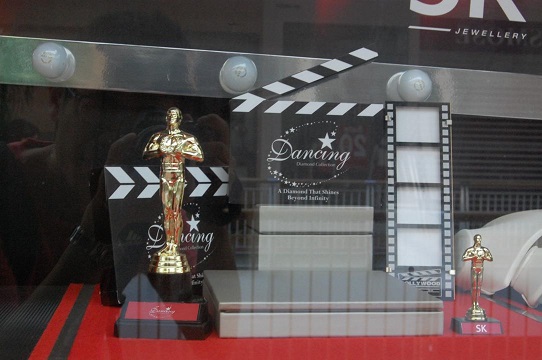}
  \vspace{-18pt}\caption{\footnotesize Input $\m I'$}
  \end{subfigure}
  \begin{subfigure}{0.138\linewidth}
  \includegraphics[width=1\textwidth]{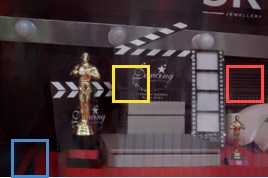}
  \vspace{-18pt}\caption{\footnotesize Initial $\m L_1'$}
  \end{subfigure}
  \begin{subfigure}{0.138\linewidth}
  \includegraphics[width=1\textwidth]{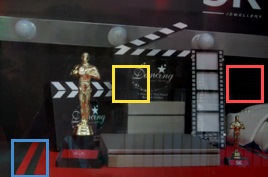}
  \vspace{-18pt}\caption{\footnotesize Final $\m L_1'$}
  \end{subfigure}
  \begin{subfigure}{0.138\linewidth}
  \includegraphics[width=1\textwidth]{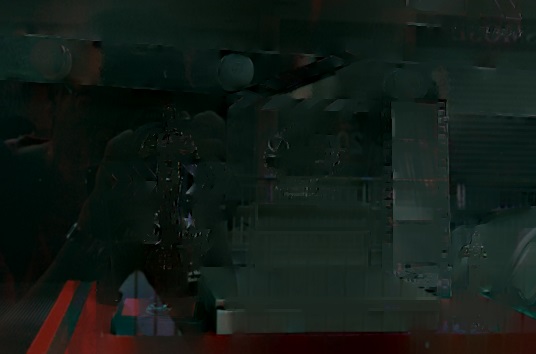}
  \vspace{-18pt}\caption{\footnotesize Initial $\m L_2'$}
  \end{subfigure}
  \begin{subfigure}{0.138\linewidth}
  \includegraphics[width=1\textwidth]{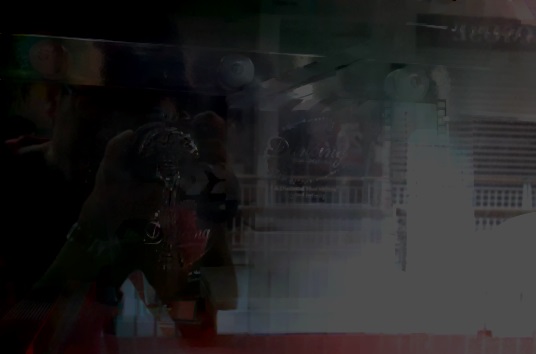}
  \vspace{-18pt}\caption{\footnotesize Final $\m L_2'$}
  \end{subfigure}
  \begin{subfigure}{0.138\linewidth}
  \includegraphics[width=1\textwidth]{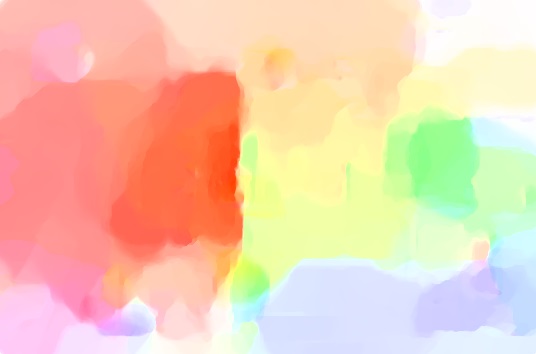}
  \vspace{-18pt}\caption{\footnotesize Initial $\m V$}
  \end{subfigure}
  \begin{subfigure}{0.138\linewidth}
  \includegraphics[width=1\textwidth]{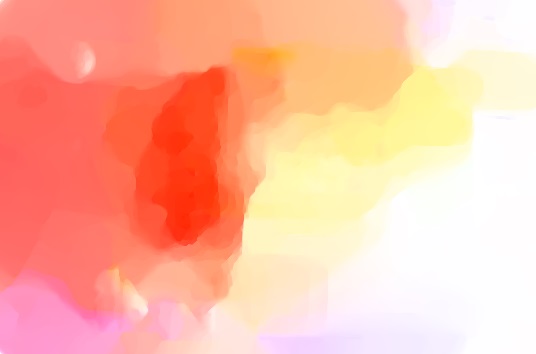}
  \vspace{-18pt}\caption{\footnotesize Final $\m V$}
  \end{subfigure}\\
  \vspace{2pt}
  \begin{subfigure}{0.243\linewidth}
  \includegraphics[width=1\textwidth]{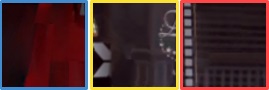}
  \vspace{-18pt}\caption{\footnotesize Close-up of initial $\m L_1$}
  \end{subfigure}~
  \begin{subfigure}{0.243\linewidth}
  \includegraphics[width=1\textwidth]{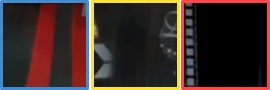}
  \vspace{-18pt}\caption{\footnotesize Close-up of final $\m L_1$}
  \end{subfigure}~~
  \begin{subfigure}{0.243\linewidth}
  \includegraphics[width=1\textwidth]{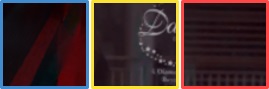}
  \vspace{-18pt}\caption{\footnotesize Close-up of initial $\m L_1'$}
  \end{subfigure}~
  \begin{subfigure}{0.243\linewidth}
  \includegraphics[width=1\textwidth]{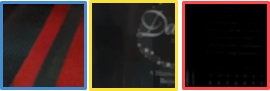}
  \vspace{-18pt}\caption{\footnotesize Close-up of final $\m L_1'$}
  \end{subfigure}

\vspace{15pt}

  \begin{subfigure}{0.138\linewidth}
  \includegraphics[width=1\textwidth]{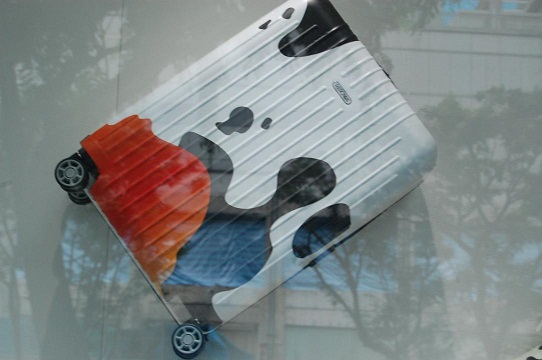}
  \vspace{-18pt}\caption{\footnotesize Input $\m I$}
  \end{subfigure}
  \begin{subfigure}{0.138\linewidth}
  \includegraphics[width=1\textwidth]{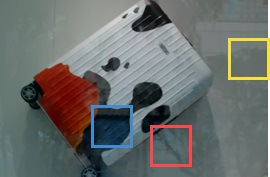}
  \vspace{-18pt}\caption{\footnotesize Initial $\m L_1$}
  \end{subfigure}
  \begin{subfigure}{0.138\linewidth}
  \includegraphics[width=1\textwidth]{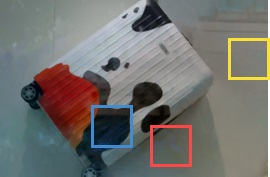}
  \vspace{-18pt}\caption{\footnotesize Final $\m L_1$}
  \end{subfigure}
  \begin{subfigure}{0.138\linewidth}
  \includegraphics[width=1\textwidth]{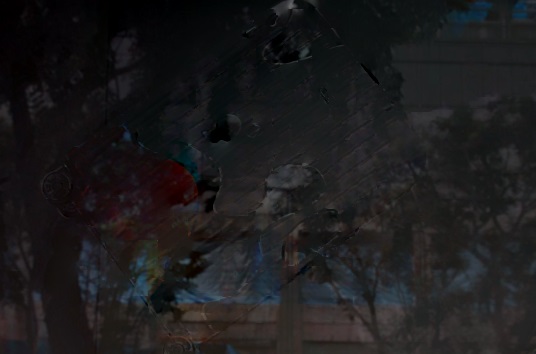}
  \vspace{-18pt}\caption{\footnotesize Initial $\m L_2$}
  \end{subfigure}
  \begin{subfigure}{0.138\linewidth}
  \includegraphics[width=1\textwidth]{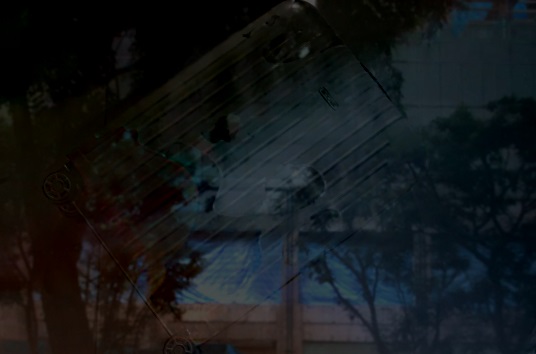}
  \vspace{-18pt}\caption{\footnotesize Final $\m L_2$}
  \end{subfigure}
  \begin{subfigure}{0.138\linewidth}
  \includegraphics[width=1\textwidth]{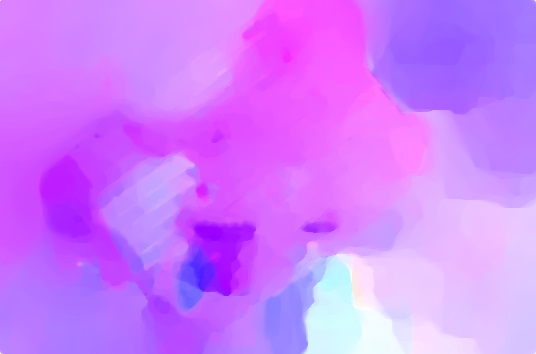}
  \vspace{-18pt}\caption{\footnotesize Initial $\m U$}
  \end{subfigure}
  \begin{subfigure}{0.138\linewidth}
  \includegraphics[width=1\textwidth]{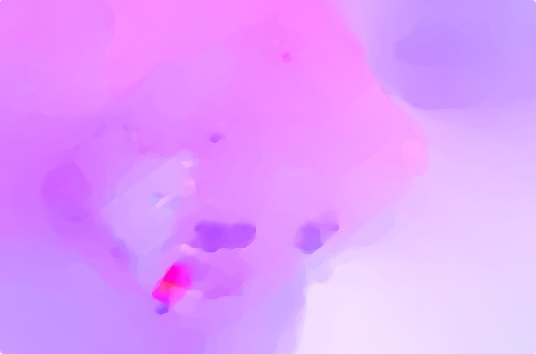}
  \vspace{-18pt}\caption{\footnotesize Final $\m U$}
  \end{subfigure}\\
  \vspace{2pt}
  \begin{subfigure}{0.138\linewidth}
  \includegraphics[width=1\textwidth]{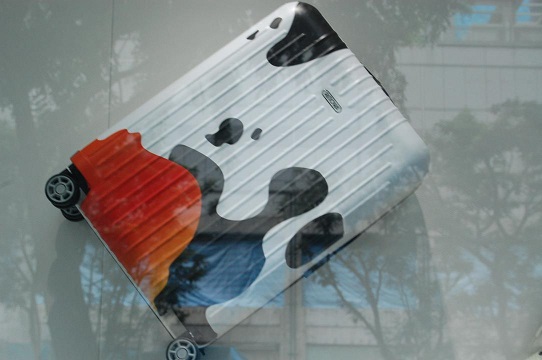}
  \vspace{-18pt}\caption{\footnotesize Input $\m I'$}
  \end{subfigure}
  \begin{subfigure}{0.138\linewidth}
  \includegraphics[width=1\textwidth]{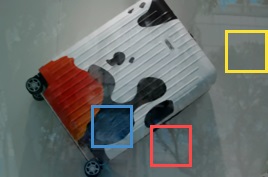}
  \vspace{-18pt}\caption{\footnotesize Initial $\m L_1'$}
  \end{subfigure}
  \begin{subfigure}{0.138\linewidth}
  \includegraphics[width=1\textwidth]{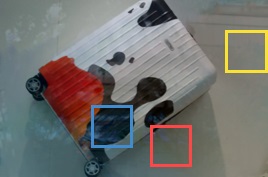}
  \vspace{-18pt}\caption{\footnotesize Final $\m L_1'$}
  \end{subfigure}
  \begin{subfigure}{0.138\linewidth}
  \includegraphics[width=1\textwidth]{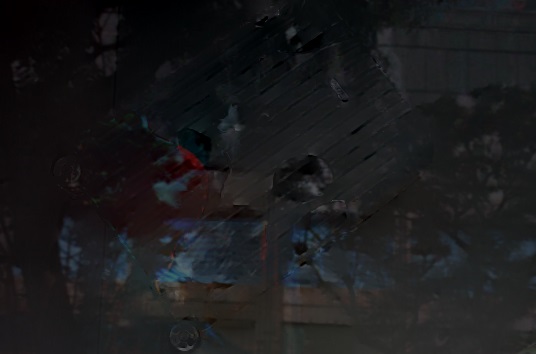}
  \vspace{-18pt}\caption{\footnotesize Initial $\m L_2'$}
  \end{subfigure}
  \begin{subfigure}{0.138\linewidth}
  \includegraphics[width=1\textwidth]{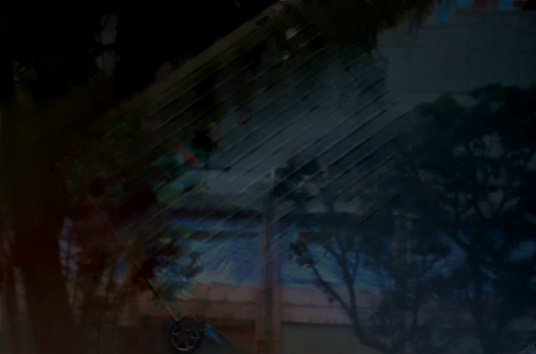}
  \vspace{-18pt}\caption{\footnotesize Final $\m L_2'$}
  \end{subfigure}
  \begin{subfigure}{0.138\linewidth}
  \includegraphics[width=1\textwidth]{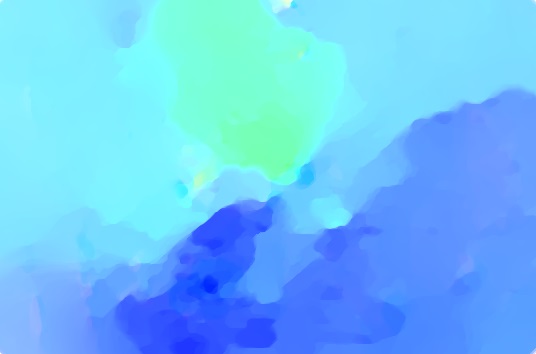}
  \vspace{-18pt}\caption{\footnotesize Initial $\m V$}
  \end{subfigure}
  \begin{subfigure}{0.138\linewidth}
  \includegraphics[width=1\textwidth]{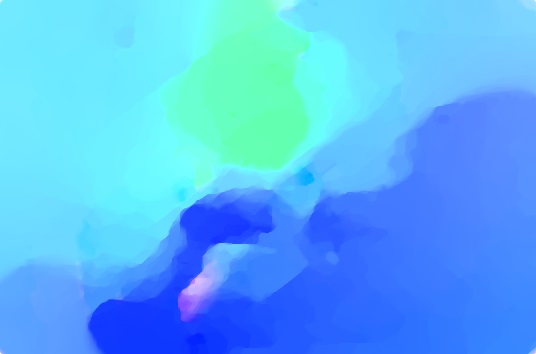}
  \vspace{-18pt}\caption{\footnotesize Final $\m V$}
  \end{subfigure}\\
  \vspace{2pt}
  \begin{subfigure}{0.243\linewidth}
  \includegraphics[width=1\textwidth]{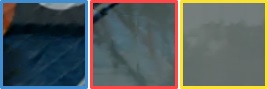}
  \vspace{-18pt}\caption{\footnotesize Close-up of initial $\m L_1$}
  \end{subfigure}~
  \begin{subfigure}{0.243\linewidth}
  \includegraphics[width=1\textwidth]{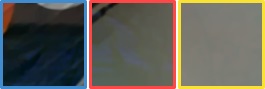}
  \vspace{-18pt}\caption{\footnotesize Close-up of final $\m L_1$}
  \end{subfigure}~~
  \begin{subfigure}{0.243\linewidth}
  \includegraphics[width=1\textwidth]{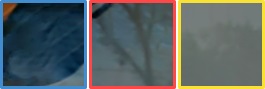}
  \vspace{-18pt}\caption{\footnotesize Close-up of initial $\m L_1'$}
  \end{subfigure}~
  \begin{subfigure}{0.243\linewidth}
  \includegraphics[width=1\textwidth]{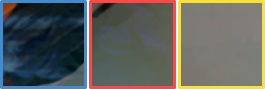}
  \vspace{-18pt}\caption{\footnotesize Close-up of final $\m L_1'$}
  \end{subfigure}
\end{center}
\vspace{-13pt}
\caption{Experimental results on real reflection images. The initial layer separations are estimated by running method of \cite{Li_brown_ICCV13} on the two input images. Visually inspected, the final optical flow fields are smoother and more consistent (see \eg the results on the back wall in the first example, and results on the floor in the second example). The corresponding warping errors are presented in Table~\ref{tab:warperr}. The close-up images in the third rows show the improvements of the reflection removal results upon the initial estimates. (\emph{\textbf{Best viewed on screen}})
}
\label{fig:double_color}
\vspace{0pt}
\end{figure*}

The initial and final results are presented in Fig.~\ref{fig:double_color}. Visually inspected, the final optical flow fields are smoother and more consistent (see e.g. the
results on the back wall in the first example, and results on the floor in the second example). As no ground truth optical flow is available, we use image warping error to quantitatively evaluate the estimated flows. The warping error for a pixel $\mathbf{x}$ in $\m L_1$ or $\m L_2$ is $\|\m L_1(\mathbf{x}\!+\!\m U(\mathbf{x}))\!-\!\m L_1'(\mathbf{x})\|_2$ or $\|\m L_2(\mathbf{x}\!+\!\m V(\mathbf{x}))\!-\!\m L_2'(\mathbf{x})\|_2$, respectively. We compute the mean warping errors for all pixels on $\m L_1$ and $\m L_2$. As shown in Table~\ref{tab:warperr}, our method has significantly reduced the warping error upon the initializations. Figure~\ref{fig:double_color} shows the improvements of the reflection removal results upon the initial estimates.

\begin{table}
  \centering
  \caption{Mean image warping errors (in gray levels) from the double-flow estimation results.}\label{tab:warperr}
  \vspace{-3pt}
  \begin{tabular}{|c|c|c|c|}
  \hline
  Image pair & Initial results & Our final results \\
  \hline
  \#1 & 6.27& \textbf{2.55} \\
  \hline
  \#2 & 3.86 & \textbf{1.49} \\
  \hline
  \end{tabular}
  \vspace{2pt}
\end{table}

\vspace{-5pt}
\paragraph{Discussion.} The dynamic foreground case with double-layer flow estimation is generally much harder than the single-flow case. This is not only because the former has more unknown variables to be solved for, but also due to the difficulties in obtaining a good initialization.
Nevertheless, our experiments show that the proposed method consistently improved the reasonable initializations given to it, for both the single-flow and double-flow cases.


\vspace{-5pt}
\paragraph{Limitation.} The proposed method is better suited for scenarios where the correlation between latent layers and their flow fields are relatively small. It will fail if both the two layers are textureless (as infinite numbers of possible motions exist satisfying the BCC constraints), or they undergo a same motion (thus the original BCC holds and only a single motion field can be extracted).

\section{Conclusions and Future Work}
This paper has defined the problem of robust optical flow estimation in the presence of possibly moving transparent or reflective layers. To our knowledge, the problem goes beyond the scope of conventional optical flow methods and was not properly investigated before.

We have presented a generalized double-layer brightness constancy condition as well as an optimization framework to solve this problem. The double-layer brightness constancy condition couples the flow fields and the brightness layers.
Encouraging experimental results of optical flow estimation and layer separation on challenging data have been obtained, even though we are using simple priors for them.
We hope that this paper can inspire future works to further address this challenging ill-posed problem.



Our current framework is based on a generative model,
which is applied uniformly to both the foreground and background layers.
In future, we plan to leverage discriminative models to exploit the differences between the two layers for better layer separation.
We also would like to explore some other optical flow priors. One possible strategy is to apply piecewise parametric motion model~\cite{ju1996skin,yang2015dense}, which provides stronger constraints than general
smoothness regularizers such as a TV, and is recently demonstrated to have advanced performances~\cite{yang2015dense}. Some other issues such as occlusion handling could also be considered.

\vspace{-5pt}
\paragraph{Acknowledgments}
{\small
H. Li's research is funded in part by Australian Research Council Grants of DP120103896, LP100100588, ARC Centre of Excellence on Robotic Vision (CE140100016) and NICTA (Data61).  Y. Dai's contribution is funded in part by ARC Grants (DE140100180, LP100100588) and National Natural Science Foundation of China (61420106007).
}

{\small
\bibliographystyle{ieee}
\bibliography{robustopticalflow_cvpr16_cr_v1.3_FINAL_authorcopy}

\begin{thebibliography}{10}\itemsep=-1pt

\bibitem{auvray2009jointmotion}
V.~Auvray, P.~Bouthemy, and J.~Li{\'e}nard.
\newblock Joint motion estimation and layer segmentation in transparent image
  sequences: application to noise reduction in {X}-ray image sequences.
\newblock {\em EURASIP Journal on Advances in Signal Processing}, pages
  19:1--19:21, 2009.

\bibitem{middlebury}
S.~Baker, D.~Scharstein, J.~Lewis, S.~Roth, M.~Black, and R.~Szeliski.
\newblock A database and evaluation methodology for optical flow.
\newblock {\em International Journal of Computer Vision (IJCV)}, 92(1):1--31,
  2011.

\bibitem{bergen1992three}
J.~R. Bergen, P.~J. Burt, R.~Hingorani, and S.~Peleg.
\newblock A three-frame algorithm for estimating two-component image motion.
\newblock {\em IEEE Transactions on Pattern Analysis and Machine Intelligence
  (TPAMI)}, (9):886--896, 1992.

\bibitem{black1996robust}
M.~J. Black and P.~Anandan.
\newblock The robust estimation of multiple motions: Parametric and
  piecewise-smooth flow fields.
\newblock {\em Computer Vision and Image Understanding (CVIU)}, (1):75--104,
  1996.

\bibitem{bredies2010total}
K.~Bredies, K.~Kunisch, and T.~Pock.
\newblock Total generalized variation.
\newblock {\em SIAM Journal on Imaging Sciences}, 3(3):492--526, 2010.

\bibitem{brox2004high}
T.~Brox, A.~Bruhn, N.~Papenberg, and J.~Weickert.
\newblock High accuracy optical flow estimation based on a theory for warping.
\newblock In {\em European Conference on Computer Vision (ECCV)}, pages 25--36,
  2004.

\bibitem{butler2012naturalistic}
D.~J. Butler, J.~Wulff, G.~B. Stanley, and M.~J. Black.
\newblock A naturalistic open source movie for optical flow evaluation.
\newblock In {\em European Conference on Computer Vision (ECCV)}, pages
  611--625, 2012.

\bibitem{chambolle2011first}
A.~Chambolle and T.~Pock.
\newblock A first-order primal-dual algorithm for convex problems with
  applications to imaging.
\newblock {\em Journal of Mathematical Imaging and Vision (JMIV)},
  40(1):120--145, 2011.

\bibitem{chartrand2008iteratively}
R.~Chartrand and W.~Yin.
\newblock Iteratively reweighted algorithms for compressive sensing.
\newblock In {\em IEEE International Conference on Acoustics, Speech and Signal
  Processing (ICASSP)}, pages 3869--3872, 2008.

\bibitem{Chen_ICCV09}
Y.~Chen, T.~Chang, C.~Zhou, and T.~Fang.
\newblock Gradient domain layer separation under independent motion.
\newblock In {\em International Conference on Computer Vision (ICCV)}, pages
  694--701, 2009.

\bibitem{darrell1993nulling}
T.~Darrell and E.~Simonecelli.
\newblock {`Nulling'} filters and the separation of transparent motions.
\newblock In {\em IEEE Conference on Computer Vision and Pattern Recognition
  (CVPR)}, pages 738--739, 1993.

\bibitem{farid1999separating}
H.~Farid and E.~H. Adelson.
\newblock Separating reflections and lighting using independent components
  analysis.
\newblock In {\em IEEE Conference on Computer Vision and Pattern Recognition
  (CVPR)}, pages 262--267, 1999.

\bibitem{gai2012blind}
K.~Gai, Z.~Shi, and C.~Zhang.
\newblock Blind separation of superimposed moving images using image
  statistics.
\newblock {\em IEEE Transactions on Pattern Analysis and Machine Intelligence
  (TPAMI)}, 34(1):19--32, 2012.

\bibitem{guo2014robust}
X.~Guo, X.~Cao, and Y.~Ma.
\newblock Robust separation of reflection from multiple images.
\newblock In {\em IEEE Conference on Computer Vision and Pattern Recognition
  (CVPR)}, pages 2195--2202, 2014.

\bibitem{irani1994computing}
M.~Irani, B.~Rousso, and S.~Peleg.
\newblock Computing occluding and transparent motions.
\newblock {\em International Journal of Computer Vision (IJCV)}, 12(1):5--16,
  1994.

\bibitem{ju1996skin}
S.~X. Ju, M.~J. Black, and A.~D. Jepson.
\newblock Skin and bones: Multi-layer, locally affine, optical flow and
  regularization with transparency.
\newblock In {\em IEEE Conference on Computer Vision and Pattern Recognition
  (CVPR)}, pages 307--314, 1996.

\bibitem{langley1992multiple}
K.~Langley, D.~Fleet, and T.~Atherton.
\newblock Multiple motions from instantaneous frequency.
\newblock In {\em IEEE Conference on Computer Vision and Pattern Recognition
  (CVPR)}, pages 846--849, 1992.

\bibitem{langley1992transparent}
K.~Langley, D.~J. Fleet, and T.~J. Atherton.
\newblock On transparent motion computation.
\newblock In {\em British Machine Vision Conference (BMVC)}, pages 247--256,
  1992.

\bibitem{levin2007user}
A.~Levin and Y.~Weiss.
\newblock User assisted separation of reflections from a single image using a
  sparsity prior.
\newblock {\em IEEE Transactions on Pattern Analysis and Machine Intelligence
  (TPAMI)}, 29(9):1647--1654, 2007.

\bibitem{levin2002learning}
A.~Levin, A.~Zomet, and Y.~Weiss.
\newblock Learning to perceive transparency from the statistics of natural
  scenes.
\newblock In {\em Advances in Neural Information Processing Systems (NIPS)},
  pages 1247--1254, 2002.

\bibitem{Li_brown_ICCV13}
Y.~Li and M.~Brown.
\newblock Exploiting reflection change for automatic reflection removal.
\newblock In {\em International Conference on Computer Vision (ICCV)}, pages
  2432--2439, 2013.

\bibitem{li2014single}
Y.~Li and M.~S. Brown.
\newblock Single image layer separation using relative smoothness.
\newblock In {\em IEEE Conference on Computer Vision and Pattern Recognition
  (CVPR)}, pages 2752--2759, 2014.

\bibitem{li2015simultaneous}
Z.~Li, P.~Tan, R.~T. Tan, D.~Zou, S.~Z. Zhou, and L.-F. Cheong.
\newblock Simultaneous video defogging and stereo reconstruction.
\newblock In {\em IEEE Conference on Computer Vision and Pattern Recognition
  (CVPR)}, pages 4988--4997, 2015.

\bibitem{pingault2002optical}
M.~Pingault and D.~Pellerin.
\newblock Optical flow constraint equation extended to transparency.
\newblock In {\em European Signal Processing Conference}, pages 1--4, 2002.

\bibitem{ramirez2006multi}
A.~Ramirez-Manzanares, M.~Rivera, P.~Kornprobst, and F.~Lauze.
\newblock Multi-valued motion fields estimation for transparent sequences with
  a variational approach.
\newblock {\em INRIA {T}echnical {R}eport}, 2006.

\bibitem{sarel2004separating}
B.~Sarel and M.~Irani.
\newblock Separating transparent layers through layer information exchange.
\newblock In {\em European Conference on Computer Vision (ECCV)}.

\bibitem{schechner2000separation}
Y.~Y. Schechner, N.~Kiryati, and R.~Basri.
\newblock Separation of transparent layers using focus.
\newblock {\em International Journal of Computer Vision (IJCV)}, 39(1):25--39,
  2000.

\bibitem{shizawa1990simultaneous}
M.~Shizawa and K.~Mase.
\newblock Simultaneous multiple optical flow estimation.
\newblock In {\em International Conference on Pattern Recognition (ICPR)},
  pages 274--278, 1990.

\bibitem{shizawa1991unified}
M.~Shizawa and K.~Maze.
\newblock Unified computational theory for motion transparency and motion
  boundaries based on eigenenergy analysis.
\newblock In {\em IEEE Conference on Computer Vision and Pattern Recognition
  (CVPR)}, pages 289--295, 1991.

\bibitem{steinbrucker2009large}
F.~Steinbr{\"u}cker, T.~Pock, and D.~Cremers.
\newblock Large displacement optical flow computation without warping.
\newblock In {\em International Conference on Computer Vision (ICCV)}, pages
  1609--1614, 2009.

\bibitem{Sun_review}
D.~Sun, S.~Roth, and M.~Black.
\newblock A quantitative analysis of current practices in optical flow
  estimation and the principles behind them.
\newblock {\em International Journal of Computer Vision (IJCV)},
  106(2):115--137, 2014.

\bibitem{sun2010layered}
D.~Sun, E.~B. Sudderth, and M.~J. Black.
\newblock Layered image motion with explicit occlusions, temporal consistency,
  and depth ordering.
\newblock In {\em Advances in Neural Information Processing Systems (NIPS)},
  pages 2226--2234.

\bibitem{szeliski2000layer}
R.~Szeliski, S.~Avidan, and P.~Anandan.
\newblock Layer extraction from multiple images containing reflections and
  transparency.
\newblock In {\em IEEE Conference on Computer Vision and Pattern Recognition
  (CVPR)}, pages 246--253, 2000.

\bibitem{szeliski1998stereo}
R.~Szeliski and P.~Golland.
\newblock Stereo matching with transparency and matting.
\newblock In {\em International Conference on Computer Vision (ICCV)}, pages
  517--524, 1998.

\bibitem{Tappen_intrinsic}
M.~F. Tappen, W.~T. Freeman, and E.~H. Adelson.
\newblock Recovering intrinsic images from a single image.
\newblock {\em IEEE Transactions on Pattern Analysis and Machine Intelligence
  (TPAMI)}, 27(9):1459--1472, 2005.

\bibitem{toro2000multiple}
J.~Toro, F.~J. Owens, and R.~Medina.
\newblock Multiple motion estimation and segmentation in transparency.
\newblock In {\em International Conference on Acoustics, Speech and Signal
  Processing (ICASSP)}, volume~6, pages 2087--2090, 2000.

\bibitem{tsin2006stereo}
Y.~Tsin, S.~B. Kang, and R.~Szeliski.
\newblock Stereo matching with linear superposition of layers.
\newblock {\em IEEE Transactions on Pattern Analysis and Machine Intelligence
  (TPAMI)}, 28(2):290--301, 2006.

\bibitem{wang1994representing}
J.~Y. Wang and E.~H. Adelson.
\newblock Representing moving images with layers.
\newblock {\em IEEE Transactions on Image Processing (TIP)}, 3(5):625--638,
  1994.

\bibitem{weiss1997smoothness}
Y.~Weiss.
\newblock Smoothness in layers: Motion segmentation using nonparametric mixture
  estimation.
\newblock In {\em IEEE Conference on Computer Vision and Pattern Recognition
  (CVPR)}, pages 520--526, 1997.

\bibitem{Weiss_intrinsic}
Y.~Weiss.
\newblock Deriving intrinsic images from image sequences.
\newblock In {\em International Conference on Computer Vision (ICCV)}, pages
  68--75, 2001.

\bibitem{Wexler_ECCV}
Y.~Wexler, A.~Fitzgibbon, and A.~Zisserman.
\newblock Bayesian estimation of layers from multiple images.
\newblock In {\em European Conference on Computer Vision (ECCV)}, pages
  487--501, 2002.

\bibitem{wulff2014modeling}
J.~Wulff and M.~J. Black.
\newblock Modeling blurred video with layers.
\newblock In {\em European Conference on Computer Vision (ECCV)}, pages
  236--252, 2014.

\bibitem{xue2015computational}
T.~Xue, M.~Rubinstein, C.~Liu, and W.~T. Freeman.
\newblock A computational approach for obstruction-free photography.
\newblock {\em ACM Transactions on Graphics (TOG)}, 34(4):79, 2015.

\bibitem{yang2015dense}
J.~Yang and H.~Li.
\newblock Dense, accurate optical flow estimation with piecewise parametric
  model.
\newblock In {\em IEEE Conference on Computer Vision and Pattern Recognition
  (CVPR)}, pages 1019--1027, 2015.

\bibitem{ye2014intrinsic}
G.~Ye, E.~Garces, Y.~Liu, Q.~Dai, and D.~Gutierrez.
\newblock Intrinsic video and applications.
\newblock {\em ACM Transactions on Graphics (TOG)}, 33(4):80, 2014.

\bibitem{yeung_CVPR08}
S.-K. Yeung, T.-P. Wu, and C.-K. Tang.
\newblock Extracting smooth and transparent layers from a single image.
\newblock In {\em IEEE Conference on Computer Vision and Pattern Recognition
  (CVPR)}, pages 1--7, 2008.

\bibitem{zach2007duality}
C.~Zach, T.~Pock, and H.~Bischof.
\newblock A duality based approach for realtime {TV-L1} optical flow.
\newblock In {\em Annual Symposium of the German Association for Pattern
  Recognition (DAGM)}, pages 214--223, 2007.

\end{thebibliography}
}

\end{document}